\RequirePackage[svgnames]{xcolor}

\documentclass[11pt,letterpaper]{mystyle}
\usepackage{tocloft}
\usepackage{setspace}
\usepackage{enumitem}

\usepackage[all]{hypcap}
\usepackage[svgnames]{xcolor}
\usepackage[comma,authoryear,compress]{natbib}
\usepackage{hyperref}[citecolor=lightblue]

\hypersetup{
    colorlinks = true,
    citecolor = {YaleBlue},
}

\usepackage{algorithm}
\usepackage{algorithmicx}
\usepackage{algpseudocode}
\usepackage{microtype}
\usepackage{graphicx}
\expandafter\def\csname ver@subfig.sty\endcsname{}
\usepackage{booktabs} %
\usepackage{float}
\usepackage{bigstrut}

\usepackage{amsmath}
\usepackage{amssymb}
\usepackage{mathtools}
\usepackage{amsthm}
\usepackage{mathrsfs}
\usepackage{nicefrac}
\usepackage{dsfont}
\usepackage{subcaption}
\usepackage{graphicx,subfig}
\usepackage{cleveref}
\usepackage{bxcoloremoji}
\usepackage{listings}
\usepackage{tcolorbox}
\tcbuselibrary{listings}

\newtcblisting{promptbox}[2][]{
    listing only,
    listing options={
        basicstyle=\ttfamily\footnotesize,
        breaklines=true,
    },
    colback=gray!10,
    colframe=YaleBlue!60, 
    title=#2,
    fonttitle=\bfseries,
    #1
}
\usepackage{enumitem}
\setlist[itemize]{leftmargin=*}
\setlist[enumerate]{leftmargin=*}
\usepackage[normalem]{ulem}

\usepackage{float}
\usepackage{multirow}
\usepackage{afterpage}

\usepackage[utf8]{inputenc} %
\usepackage[T1]{fontenc}    %
\usepackage{hyperref}       %
\usepackage{url}            %
\usepackage{booktabs}       %
\usepackage{amsfonts}       %
\usepackage{nicefrac}       %
\usepackage{microtype}      %
\usepackage{graphicx}
\usepackage{subcaption} 
\usepackage{amssymb}
\usepackage{fdsymbol}
\usepackage{wrapfig}
\usepackage{lipsum}
\usepackage{stackengine}
\usepackage[font=small,labelfont=bf]{caption}
\usepackage{color}
\usepackage{adjustbox}

\usepackage{rotating}
\usepackage{makecell}
\usepackage{listings}  
\newcommand{\x}{\mathbf{x}}

\definecolor{blanchedalmond}{rgb}{1.0, 0.92, 0.8}
\definecolor{carmine}{rgb}{0.59, 0.0, 0.09}
\definecolor{lightblue}{rgb}{0.22,0.45,0.70}%

\renewcommand{\mathbf}{\boldsymbol}

\makeatletter
\def\Ddots{\mathinner{\mkern1mu\raise\p@
\vbox{\kern7\p@\hbox{.}}\mkern2mu
\raise4\p@\hbox{.}\mkern2mu\raise7\p@\hbox{.}\mkern1mu}}
\makeatother

\definecolor{amaranth}{rgb}{0.9, 0.17, 0.31}
\definecolor{antiquebrass}{rgb}{0.8, 0.58, 0.46}
\definecolor{antiquefuchsia}{rgb}{0.57, 0.36, 0.51}
\definecolor{chromeyellow}{rgb}{0.31, 0.47, 0.26}

\newtcolorbox{AIbox}[2][]{aibox,title=#2,#1}
\definecolor{lightblue}{rgb}{0.22,0.45,0.70}%
\definecolor{Gray}{gray}{0.95}
\definecolor{Cornsilk}{rgb}{1.0, 0.97, 0.86}

\usepackage{amsmath}

\usepackage{pdfrender}

\definecolor{Gray}{gray}{0.95}
\newcolumntype{a}{>{\columncolor{Gray}}c}

\usepackage{skak}
\usepackage{bbm}
\usepackage{lipsum}
\usepackage{listings}
\usepackage{caption}
\usepackage{fancyhdr}

\definecolor{pink}{HTML}{fc6c85}

\definecolor{customblue}{HTML}{286dc0}

\definecolor{customred}{HTML}{CC0000}

\definecolor{customyellow}{HTML}{ffd55a}

\definecolor{customgrey}{HTML}{978d85}

\newtcolorbox{blueBox}[1][]{
  colback=YaleBlue!12!white,  
  colframe=YaleBlue!85,   
  boxrule=1pt,
  arc=3pt,
  floatplacement=floating,
  title=\centering #1
}

\newtcolorbox{yellowBox}[1][]{
  colback=customyellow!10!white,
  colframe=customyellow,
  floatplacement=floating,
  title=\centering #1
}

\newtcolorbox{promptBox}[1][]{
  colback=YaleBlue!12!white,  
  colframe=YaleBlue!85,      
  boxrule=0.8pt,
  arc=3pt,
  floatplacement=floating,
  title=\centering #1
}

\newtcolorbox{wronganswer}[1][]{
    enhanced,
    breakable,
    colframe=customred!65,
    colback=customred!6!white,
    sharp corners,
    boxsep=0pt,
    left=5pt,
    right=5pt,
    top=6pt,
    bottom=6pt,
    boxrule=0pt,
    leftrule=4pt,
    #1
}

\newtcolorbox{correctanswer}[1][]{
    enhanced,
    breakable,
    colframe=OliveGreen,
    colback=OliveGreen!10!white,
    sharp corners,
    boxsep=0pt,
    left=5pt,
    right=5pt,
    top=6pt,
    bottom=6pt,
    boxrule=0pt,
    leftrule=4pt,
    #1
}

\usepackage{fontawesome}
\DeclareFontShape{T1}{cmss}{m}{up}{<->ssub*cmss/m/n}{}

\title{Length Value Model: Scalable Value Pretraining for Token-Level Length Modeling}
\runningtitle{Length Value Model: Scalable Value Pretraining for Token-Level Length Modeling}

\newcommand{\ucsb}{1}
\newcommand{\cmu}{2}
\newcommand{\wisc}{3}
\newcommand{\lmsys}{4}
\newcommand{\apple}{5}

\author[\ucsb]{Zhen Zhang}
\author[\cmu,\lmsys]{Changyi Yang}
\author[\lmsys]{Zijie Xia}
\author[\apple]{Zhen Yang}
\author[\ucsb]{Chengzhi Liu}
\author[\ucsb]{Zhaotiao Weng}
\author[\ucsb]{Yepeng Liu}
\author[\ucsb]{Haobo Chen}
\author[\wisc,\lmsys]{Jin Pan}
\author[\lmsys]{Chenyang Zhao}
\author[\ucsb]{Yuheng Bu}
\author[\apple]{Alkesh Patel}
\author[\apple]{Zhe Gan}
\author[\ucsb]{Xin Eric Wang}

\affil[\ucsb]{University of California, Santa Barbara}
\affil[\cmu]{Carnegie Mellon University}
\affil[\wisc]{University of Wisconsin–Madison}
\affil[\lmsys]{LMSYS Org}
\affil[\apple]{Apple Inc.}

\begin{document} 

\begin{abstract}
\vspace{-9pt}
\footnotesize
Tokens are the fundamental units of computation in modern autoregressive models, and generation length directly influences both inference cost and reasoning performance. Despite its importance, existing approaches model length primarily at the coarse sequence level. We introduce the Length Value Model (LenVM), a token-level framework that estimates the remaining generation length at every decoding step. By formulating length modeling as a value estimation problem and assigning a constant negative reward to each generated token, LenVM predicts a bounded, discounted return that is a monotone proxy for the remaining generation horizon. This value formulation provides annotation-free, dense, unbiased, and scalable supervision.
Experiments on LLMs and VLMs show that LenVM supports exact control, continuous performance--efficiency steering, length prediction, and interpretation. On LIFEBench-token, it raises the exact-length score of Qwen2.5-7B-Instruct from $30.9$ to $64.8$ in one pass; combined with eight-round LCG, it further raises the score from $71.3$ to $83.6$ while reducing average absolute deviation from $14\%$ to $6\%$. On GSM8K near $200$ tokens, LenVM retains about $63\%$ Pass@1 versus $6\%$ for a hard token budget, and it outperforms budget-aware prompting and EOS calibration at matched lengths on MATH500. Prompt-boundary prediction improves consistently with scale, reaching MREs of $9.8\%$, $14.9\%$, and $17.1\%$ on math, code, and instruction following at 32B; against entropy-guided predictors, LenVM reduces prompt MRE from $32.56$ to $26.90$ and progressive MRE from $26.71$ to $15.10$. Its token-level values also identify pivots between longer and shorter trajectories.
Results demonstrate that LenVM supports a broad range of applications, including length control, prediction, and interpretation of generation dynamics. They suggest that generation length can be effectively modeled as a token-level value signal, highlighting the potential of LenVM as a general framework for length modeling and as a length-specific value signal that supports future RL training.  
\vspace{2.5pt}

\parbox{\linewidth}{
\textbf{Correspondence:} \href{mailto:zhen_zhang@ucsb.edu}{zhen\_zhang@ucsb.edu}, \href{mailto:ericxwang@ucsb.edu}{ericxwang@ucsb.edu}\\
\textbf{Project Page:} \href{https://github.com/eric-ai-lab/Length-Value-Model}{https://github.com/eric-ai-lab/Length-Value-Model} \parbox[t]{0.5\linewidth}{\raggedright\url{}}
}
\end{abstract}

{\hbadness=10000\maketitle}

\begin{figure}[h]
    \centering
    \vspace{-5pt}
    \includegraphics[width=0.93\linewidth]{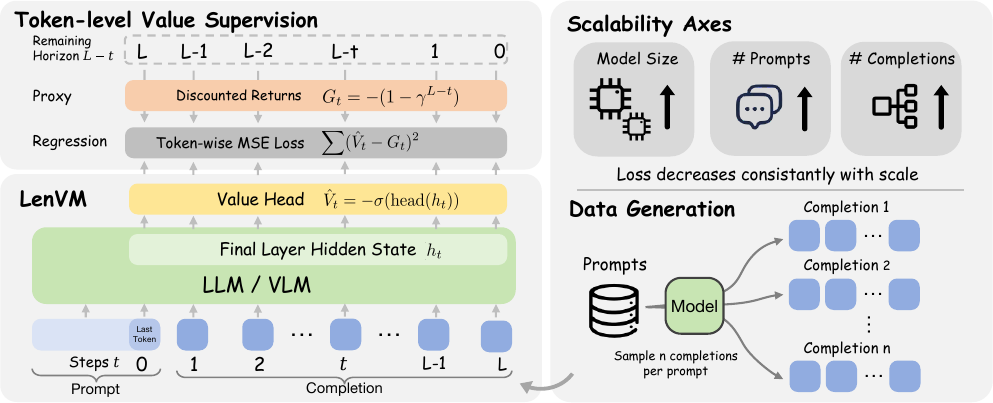}
    \vspace{-5pt}
    \caption{\textbf{Length Value Model (LenVM) architecture and Pipeline.} At each decoding step, LenVM attaches a scalar value head to the final-layer hidden state of an LLM or VLM to predict the value $V_t \in (-1, 0)$. LenVM enables annotation-free and scalable value pretraining across model size, prompt count, and number of completions.}
    \label{fig:lenvm_arch}
\end{figure}

\renewcommand{\cftsecfont}{\small}
\renewcommand{\cftsubsecfont}{\small}
\renewcommand{\cftsecpagefont}{\small}
\renewcommand{\cftsubsecpagefont}{\small}

\setlength{\cftbeforesecskip}{3pt}
\setlength{\cftbeforesubsecskip}{1pt}

\footnotesize \tableofcontents

\newpage

\section{Introduction}
\label{sec:introduction}

Token generation is the core computational process in modern AI systems. Large language models (LLMs), vision-language models (VLMs), and specialized models for code, mathematics, and agentic tasks all produce outputs sequentially. The number of generated tokens therefore forms a fundamental unit of inference-time computation. It influences both \emph{performance}, since additional tokens can enable more reasoning in challenging settings~\citep{Snell2024ScalingLT}, and \emph{cost}, since each token incurs compute, KV-cache memory, latency, and energy~\citep{Kwon2023EfficientMM,Pope2022EfficientlyST}. Despite its importance, existing approaches to length modeling and control remain coarse-grained. Sequence-level penalties during training~\citep{Team2025KimiKS}, prompt-based instructions at inference~\citep{Xiao2026CanLT}, and pre-decode predictors~\citep{Zhang2025AdaptThinkRM} all operate at the level of whole sequences or fixed pre-generation decisions, rather than the token-level dynamics of generation itself. What is missing is a token-level model of remaining generation length.

To address this gap, we propose the \textbf{Length Value Model (LenVM)}, a token-level formulation of generation length modeling. 
At each decoding step, LenVM predicts a scalar value for the current decoding state. Conceptually, the quantity of interest is the remaining generation length, which depends on how much future generation is still to be realized. We model this quantity through a value formulation: by assigning a constant negative reward to each generated token and discounting future steps by $\gamma$, we define a discounted return that serves as a bounded, monotone proxy for the remaining generation horizon. LenVM is trained to predict this return, yielding a principled reduction from token-level length modeling to value estimation and placing the problem in a standard value-function framework. In this view, LenVM evaluates a decoding state by its remaining distance to completion rather than by task utility. This construction also maps highly variable raw lengths into a bounded target space, improving conditioning when generation lengths range from a few tokens to over 32k. The resulting value signal can be used at inference time for length estimation and control, and at training time as a length-aware value baseline.

A central advantage of LenVM is that it naturally supports \textbf{scalable value pretraining}. The resulting supervision signal has four favorable properties. First, it is \textbf{annotation-free}: training targets are constructed automatically from sampled completions, without any additional human labeling or reward modeling. Second, it is \textbf{dense}: every token position contributes a target, rather than only one target per completed response. Third, it is \textbf{unbiased} for the policy-conditioned value target: under a fixed rollout policy, the realized return is an unbiased Monte Carlo sample of the target value and introduces no additional annotation or reward-model noise. Fourth, it is \textbf{scalable}: supervision grows naturally with both prompt count and the number of completions sampled per prompt, while remaining cheap to construct and well matched to increasing model size. Together, these properties make LenVM well suited to large-scale value pretraining, enabling high-quality token-level value supervision without the annotation bottlenecks of conventional reward and value modeling.

We evaluate whether LenVM learns a meaningful token-level signal of remaining generation length. At inference time, LenVM exposes a smooth \textbf{performance--efficiency trade-off}: exponentially tilting the token distribution with predicted values traces a Pareto frontier between response quality and generation length. Comparisons with budget-aware prompting, EOS calibration, and a length-penalized RLVR model further test this trade-off at matched generation lengths (\S~\ref{sec:tradeoff}). The same signal also supports explicit \textbf{length control}, where LenVM-guided decoding improves adherence to Equal To, At Most, and At Least constraints and complements iterative black-box control on LIFEBench-token~\citep{zhang2025lifebench} (\S~\ref{sec:length_controlled_generation}). In addition, LenVM provides useful \textbf{length prediction} from both the prompt boundary and progressive prefix states, across math, code, and instruction-following domains and against specialized entropy-guided predictors (\S~\ref{sec:first_token_pred}). Its token-level values further offer an interpretable view of generation dynamics, revealing where trajectories shift between shorter and longer reasoning regimes (\S~\ref{sec:deep_thinking_token}). Beyond inference, LenVM admits a natural interpretation as a \textbf{length-specific value baseline} in PPO-style RL~\citep{Schulman2017ProximalPO}, while its value-learning objective improves consistently with model and data scale (\S~\ref{sec:scalability}). Finally, packed scoring and entropy skipping substantially reduce \textbf{inference overhead} (\S~\ref{sec:inference_cost}). Together, these results establish LenVM as a principled and scalable framework for token-level length modeling, while leaving empirical RL fine-tuning to future work.
\section{Related Work}
\label{sec:related_work}

\subsection{Length-Controlled Generation}
Length control has been explored through prompting, fine-tuning, and decoding-time constraints.
Prompt-based methods can be attractive for black-box models: for example, Xie and Lee propose a one-shot countdown prompt that improves strict compliance without changing model weights~\citep{Xie2025PromptBasedOE}.
Recent prompt-only methods also incorporate explicit structure and counting to regulate length without retraining; Plan-and-Write uses a plan-first scaffold and word-counting guidance to improve adherence in summarization tasks~\citep{Akinfaderin2025PlanandWriteSL}.
For black-box settings where fine-tuning is not possible, \citet{Gu2024LengthCG} propose an iterative sampling framework that combines Metropolis--Hastings with importance-sampling acceleration to achieve length-constrained generation without modifying model parameters.
More broadly, constrained generation has a long history of MCMC-style inference-time methods; CGMH applies Metropolis--Hastings moves to satisfy lexical constraints without parallel training data~\citep{miao2018cgmhconstrainedsentencegeneration}, and recent work revisits MCMC constrained sampling with stronger distributional guarantees~\citep{Gonzalez2025ConstrainedSF}.
Training-based approaches can yield more precise control by modifying the model's internal length representation; \citet{Butcher2024PreciseLC} introduce length-difference positional encoding (LDPE) and fine-tuning to encourage coherent termination near the target length , while \citet{Xie2026PredictingLO} propose Hansel, which uses periodically outputted hidden special tokens to track the remaining target length during finetuning.
In contrast to approaches that embed length information into the generator, our method trains a \emph{separate} token-level horizon estimator and uses it as a decoding-time control signal, enabling length control without retraining the base LLM.

\subsection{Output Length Prediction}
Predicting completion length is important for both serving efficiency and stochastic sampling, and can be approached either statically (before decoding) or progressively (during decoding).
Recent work shows that hidden representations and entropy signals can provide accurate, low-overhead length prediction; in particular, \citet{Xie2026PredictingLO} propose entropy-guided token pooling for static prediction and progressive length prediction (PLP) for online remaining-length estimation .
Beyond entropy-guided methods, \citet{piotrowski-etal-2025-will} study forecasting remaining output length from frozen layerwise hidden states, exploring aggregation and graph-based regressors .
In code generation, horizon-length prediction has also been used to improve fill-in-the-middle planning: \citet{ding2025planningawarecodeinfillinghorizonlength} propose a horizon-length prediction objective to better align infills with distant right context.
Relatedly, \citet{xiao2026llmstrackoutputlength} examine whether LLMs can self-track output length and propose a dynamic feedback mechanism that adjusts generation online to meet length targets without retraining.
Our LVM is also an online predictor, but is trained as a standalone value model with bounded regression targets and is designed to provide a stable per-token signal across diverse task families and length scales.
Variable output lengths complicate batching and can waste compute through padding and fragmentation; length prediction can improve schedulers and load balancing.
\citet{zheng2023responselengthperceptionsequence} study response length perception and sequence scheduling to improve inference throughput.

\subsection{RL Framing and Reward Shaping of Generation Length}
Our formulation views length regularization as a dense per-step penalty (i.e., negative reward) with an associated discounted return, making remaining horizon a value-function quantity.
This makes LVM naturally compatible with PPO-style training as a length-specific value baseline and advantage component (Section~\ref{sec:rl_reward_signal}), while in this paper we focus on inference-time decoding control and analysis.
More broadly, recent RLHF work argues for value-centric formulations: DVPO pretrains a global value model on preference data and uses it as a frozen critic for policy optimization~\citep{huang2026pretrainvaluerewarddecoupled}, and $V_0$ learns a generalist value estimator at the initial prompt for policy-agnostic routing and resource allocation~\citep{zhang2026v0generalistvaluemodel}.
Recent work on efficient reasoning also studies how to adapt length penalties during RL to balance accuracy and conciseness~\citep{xiang2025justthinkingefficientreasoning,liu2025dlerdoinglengthpenalty,li2025leashadaptivelengthpenalty}. VAPO further targets stable and efficient value-based RL for long-CoT reasoning by addressing value bias, heterogeneous sequence lengths, and sparse rewards~\citep{yue2025vapoefficientreliablereinforcement}.


\section{Length Value Model}
\label{sec:length_value_model}

We propose the \textbf{Length Value Model} (LenVM), a token-level value function over decoding states for modeling generation length.
As shown in Figure~\ref{fig:lenvm_arch}, LenVM treats autoregressive generation as an episodic process and models remaining generation length through a discounted return induced by a constant per-token cost.
Rather than predicting raw remaining length directly, it estimates a bounded and monotone transform of the remaining generation horizon.
Starting from the last prompt token, LenVM outputs a scalar value at each decoding step.
We first define the return construction (\S\ref{sec:reward_return}), then summarize its key properties, and finally describe the prediction head (\S\ref{sec:lenvm_arch}) and training objective (\S\ref{sec:training_objective}).

\subsection{Modeling Length as a Discounted Return}
\label{sec:reward_return}

Let $s_t$ denote the decoding state after the prompt and the first $t$ generated tokens, with $s_0$ corresponding to the last prompt token.
For a sampled completion of generated length $L$, decoding steps are indexed by $t \in \{0,\dots,L\}$, where $t=L$ is the EOS step.
Our goal is to estimate, from each prefix state $s_t$, how much generation horizon remains under the rollout policy that produced the sampled trajectory.

We assign a constant negative reward at each non-terminal decoding step:
\begin{equation}
r_t = -(1-\gamma), \qquad t=0,\dots,L-1,
\end{equation}
where $\gamma \in (0,1)$ is the discount factor, and define $r_L = 0$.
The factor $(1-\gamma)$ is introduced purely for normalization, so the resulting return lies in a fixed range compatible with a sigmoid-bounded output.

The discounted return from step $t$ is
\begin{equation}
G_t \triangleq \sum_{i=0}^{L-t} \gamma^i r_{t+i}
= -(1-\gamma^{L-t}).
\end{equation}
Thus, for each non-terminal step, $G_t \in (-1,0)$ and is a strictly monotone function of the remaining generation length $L-t$:
states closer to termination have values nearer to $0$, while states with longer continuations have values closer to $-1$.
Compared with raw length, this bounded transform preserves ordering while improving target scale and conditioning.
It also satisfies the Bellman recursion
\begin{equation}
G_t = r_t + \gamma G_{t+1},
\end{equation}
placing token-level length modeling in a standard value-learning framework.

\paragraph{Key properties of the target.}
The return construction has three useful properties.
First, it is \textbf{bounded}: $G_t \in (-1,0)$ for all non-terminal steps, avoiding the large dynamic range of raw remaining length.
Second, it is \textbf{monotone}: the ordering over remaining horizons is preserved exactly, so shorter continuations always map to values closer to $0$.
Third, it is \textbf{Bellman-consistent}: the target obeys a one-step recursion, which places token-level length modeling in the same formal framework as standard discounted value estimation.

The discount factor $\gamma$ controls how aggressively long horizons are compressed.
Larger $\gamma$ preserves more resolution over long continuations, while smaller $\gamma$ concentrates resolution near termination.
Equivalently, $\gamma$ determines the logarithmic scale on which remaining length is represented.

For a fixed model-induced decoding policy $\pi$, the corresponding length value function is
\begin{equation}
V^\pi(s_t) = \mathbb{E}_{\pi}[G_t \mid s_t].
\end{equation}
In practice, we sample trajectories from a fixed rollout policy and regress to their realized returns as Monte Carlo supervision.
When applied to a realized return, the transform can be inverted exactly to recover the realized remaining token count:
\begin{equation}
l = \frac{\log(1+G_t)}{\log \gamma}.
\end{equation}
When applied to a predicted conditional mean value, however, this inverse yields a proxy-derived length estimate rather than, in general, the conditional expectation of raw remaining length. We discuss this distinction in Appendix~\ref{app:length_underestimation}.

This formulation is (1) \textbf{annotation-free}, since targets are computed directly from observed completion lengths; (2) \textbf{dense}, since every non-terminal step contributes a regression target; (3) \textbf{unbiased} for the policy-conditioned value target, since under a fixed rollout policy the realized return is an unbiased Monte Carlo sample of $V^\pi(s_t)$ and is computed deterministically from the sampled completion, introducing no additional annotation or reward-model noise; and (4) \textbf{scalable}, since multiple completions can be sampled per prompt to automatically construct many supervised trajectories.

\subsection{LenVM Head}
\label{sec:lenvm_arch}

LenVM is built on top of LLMs or VLMs by attaching a scalar value head to the final-layer hidden state at each decoding step:
\begin{equation}
z_t = \mathrm{head}(h_t).
\end{equation}
In this work, we instantiate $\mathrm{head}(\cdot)$ as a two-layer MLP with SiLU activation,
\begin{equation}
z_t = W_2 \,\mathrm{SiLU}(W_1 h_t + b_1) + b_2,
\end{equation}
followed by
\begin{equation}
V_\theta(s_t) = -\sigma(z_t),
\end{equation}
which ensures $V_\theta(s_t) \in (-1,0)$.

\subsection{Training Objective}
\label{sec:training_objective}

For a minibatch of $N$ prompt-completion trajectories, where the $n$-th trajectory has generated length $L^{(n)}$, we optimize the token-averaged mean squared error
\begin{equation}
\mathcal{L}_{\mathrm{len}}
=
\frac{
\sum_{n=1}^{N}\sum_{t=0}^{L^{(n)}-1}
\bigl(V_\theta(s_t^{(n)}) - G_t^{(n)}\bigr)^2
}{
\sum_{n=1}^{N} L^{(n)}
},
\end{equation}
where
\begin{equation}
G_t^{(n)} = -(1-\gamma^{L^{(n)}-t}).
\end{equation}
Here the trajectories are sampled from a fixed generator checkpoint and decoding policy, and $G_t^{(n)}$ is computed exactly from each realized completion.
This objective therefore corresponds to Monte Carlo regression with dense token-level supervision over the rollout state distribution induced by that policy.

More generally, one may consider weighted variants of the same objective, where each token-level squared error is multiplied by a weight. In our setting, we avoid weights that depend on future rollout outcomes relative to the current decoding step, since such weights change the effective regression objective. Trajectory-level averaging (token-mean then sequence-mean) is one such case, as it induces a per-token weight inversely proportional to the realized completion length. We discuss this issue in Appendix~\ref{app:future_dependent_weighting}.

\paragraph{Relation to GAE.}
LenVM also admits GAE~\citep{Schulman2015HighDimensionalCC}.
With
$\delta_t = r_t + \gamma V_\theta(s_{t+1}) - V_\theta(s_t)$,
one can construct bootstrapped targets from temporal-difference residuals.
This provides an optional bootstrapped variant of the same value-learning problem.
In our experiments, however, $\lambda=1$ performs best.
This is consistent with the task structure: supervision is dense and directly recoverable from completed sequences, so full-return regression is already available without additional bootstrapping assumptions.

\section{Experiments: Validating LenVM as a Token-Level Length Signal}
\label{sec:lvm_validation}

In this section, we evaluate at inference time how well LenVM captures the token-level signal of remaining generation length.
Our experiments ask three questions:
\textbf{(i)} Can LenVM support inference-time control without modifying the base generator?
\textbf{(ii)} Can it predict generation length from the prompt boundary?
\textbf{(iii)} Does its value-learning objective scale with model and data?
We answer these questions through four evaluations:
\textbf{(1)~length-controlled generation} (\S\ref{sec:length_controlled_generation}),
\textbf{(2)~performance--efficiency trade-off} (\S\ref{sec:tradeoff}),
\textbf{(3)~generation length prediction} (\S\ref{sec:first_token_pred}), and
\textbf{(4)~scalability of LenVM} (\S\ref{sec:scalability}).

Applying LenVM adds forward passes; Section~\ref{sec:inference_cost} reports the resulting latency and an entropy-skipping optimization.

\subsection{Experimental Setup}
\label{sec:exp_setup}

\begin{table*}[!t]
  \centering
  \caption{\textbf{Datasets used to train general LenVMs.}}
  \label{tab:exp_domains}
  \begin{adjustbox}{max width=\textwidth}
  \begin{tabular}{lll}
    \toprule
    Domain & Dataset & Scale \\
    \midrule
    Code & OpenCodeReasoning-2~\citep{ahmad2025opencodereasoning} (Python) & 1.42M \\
    Instruction Following & WildChat~\citep{zhao2024wildchat} & 529k \\
    Math & DeepMath-103K~\citep{he2025deepmath} & 103k \\
    \bottomrule
  \end{tabular}
  \end{adjustbox}
\end{table*}

\paragraph{General LenVM.} We train LenVMs on a multi-domain mixture of math, code, and instruction-following data rather than training separate domain-specific models.
Table~\ref{tab:exp_domains} summarizes the training mixture.
For each prompt, we sample multiple completions and convert them into dense per-token regression targets using the discounted return in \S\ref{sec:length_value_model}.

\paragraph{Models.} For Qwen2.5 experiments (both LLMs and VLMs), we initialize LenVM from the \textbf{Qwen2.5-Instruct} family~\citep{Yang2024Qwen25TR}.
For Qwen3 experiments, we initialize LenVM from \textbf{Qwen3-Base} models~\citep{Yang2025Qwen3TR}.

Further experimental details appear in Appendix~\ref{sec:add_exp_details}.

\subsection{Application 1: Length-Controlled Generation}
\label{sec:length_controlled_generation}

We first test inference-time control on \textbf{LIFEBench-token}, our token-based adaptation of LIFEBench~\citep{zhang2025lifebench}. It contains 360 bilingual instances, three constraints (\emph{Equal To}, \emph{At Most}, and \emph{At Least}), and targets from 32 to 1024 tokens. We retain the tasks, target grid, and metrics but replace word or character counts with model-specific token counts, so scores are not directly comparable with the original leaderboard. We report \textbf{Length Score (LS$\uparrow$)} for all settings and \textbf{Length Deviation (LD$\downarrow$)} for \emph{Equal To}; Appendix~\ref{app:lifebench_details} gives full definitions.

\paragraph{Exact-control baselines.}
We compare with Prompt Best-of-$N$ (sample and select), CAPEL (prompted countdown)~\citep{Xie2025PromptBasedOE}, Dynamic Feedback (token-count feedback between segments)~\citep{Xiao2026CanLT}, and LCG (iterative revision with length and quality acceptance)~\citep{Gu2024LengthCG}. LCG + LenVM uses guided decoding within each proposal. One call is one generation or refinement pass; Appendix~\ref{sec:add_exp_details} gives the implementations.

\paragraph{Hard-constraint decoding with LenVM.}
At each step, we first restrict the vocabulary to a candidate set $\mathcal{V}_t$ using standard truncation strategies, then score each candidate token with LenVM on the next state $\hat{v}(x)=\hat{v}_\phi(s_t \oplus x)$.
We then select tokens using:
\begin{itemize}
    \item \textbf{Equal To}: First convert the remaining target length $L-t$ at into value space $v^*_t$ and select $\arg\min_{x\in\mathcal{V}_t} |\hat{v}(x)-v_t^\star|$, favoring candidates whose predicted value is closest to the target $v_t^\star$.
    \item \textbf{At Least}: $\arg\min_{x\in\mathcal{V}_t} \hat{v}(x)$, favoring more negative values and hence longer continuations.
    \item \textbf{At Most}: $\arg\max_{x\in\mathcal{V}_t} \hat{v}(x)$, favoring values closer to $0$ and hence earlier termination.
\end{itemize}

\begin{table}[!t]
  \centering
  \caption{\textbf{Length control on LIFEBench-token.} LS measures adherence in $[0,100]$ (higher is better; 100 is perfect). Equal LD is mean absolute relative deviation (lower is better). N/A marks methods evaluated only for exact control.}
  \label{tab:lifebench_length_control}
  \small
  \setlength{\tabcolsep}{6pt}
  \begin{tabular}{lcrrrr}
    \toprule
    Model or method & Calls & Equal LD $\downarrow$ & Equal LS $\uparrow$ & At Most LS $\uparrow$ & At Least LS $\uparrow$ \\
    \midrule
    \multicolumn{6}{l}{\textit{Closed-source benchmark models}} \\
    GPT-4o & 1 & 74\% & 35.5 & 77.9 & 98.5 \\
    GPT-5.4 & 1 & 135\% & 37.4 & 65.4 & 98.9 \\
    GPT-5.4-thinking & 1 & 131\% & 47.8 & 72.7 & 98.9 \\
    Claude-Sonnet-4-6 & 1 & 105\% & 34.1 & 62.9 & 100.0 \\
    Claude-Sonnet-4-6-thinking & 1 & 124\% & 51.3 & 69.3 & 100.0 \\
    Claude-Opus-4-6 & 1 & 66\% & 35.5 & 51.5 & 100.0 \\
    Claude-Opus-4-6-thinking & 1 & 87\% & \underline{53.2} & 67.4 & 100.0 \\
    Gemini-3-Flash-Preview & 1 & 123\% & 40.3 & 57.3 & 99.6 \\
    Gemini-3.1-Pro-Preview & 1 & 91\% & 49.3 & 70.7 & 100.0 \\
    \midrule
    \multicolumn{6}{l}{\textit{Open benchmark models, LD}} \\
    Qwen2.5-3B-Instruct & 1 & 83\% & 25.6 & 92.1 & 94.6 \\
    \quad + LenVM (1.5B) & 1 & 56\% & \textbf{62.6} & 93.0 & 93.1 \\
    Qwen2.5-7B-Instruct & 1 & 71\% & 30.9 & 98.5 & 89.1 \\
    \quad + LenVM (1.5B) & 1 & 44\% & \textbf{64.8} & 96.1 & 99.5 \\
    Qwen3-30B-A3B-Instruct & 1 & 90\% & 36.8 & 87.0 & 99.3 \\
    \quad + LenVM (1.7B) & 1 & 57\% & \textbf{67.2} & 99.4 & 99.8 \\
    \midrule
    \multicolumn{6}{l}{\textit{Exact-control methods on Qwen2.5-7B-Instruct}} \\
    CAPEL & 1 & 141\% & 28.8 & N/A & N/A \\
    Prompt Best-of-8 & 8 & 29\% & 58.1 & N/A & N/A \\
    LenVM & 1 & 44\% & 64.8 & N/A & N/A \\
    Dynamic Feedback & 8 & 16\% & 66.6 & N/A & N/A \\
    LCG & 8 & 14\% & 71.3 & N/A & N/A \\
    LCG + LenVM & 8 & \textbf{6\%} & \textbf{83.6} & N/A & N/A \\
    \bottomrule
  \end{tabular}
\end{table}

\paragraph{Results.} Table~\ref{tab:lifebench_length_control} shows that LenVM substantially improves constraint following, especially in the \emph{Equal To} setting.
For example, on Qwen2.5-7B-Instruct, LenVM improves Length Score from 30.9 to 64.8 and reduces deviation from 71\% to 44\%.
The closed-source models obtain Equal-To deviations between $66\%$ and $135\%$ and scores below $54$ in this token-based setting. Against prompt Best-of-$N$, CAPEL~\citep{Xie2025PromptBasedOE}, Dynamic Feedback~\citep{Xiao2026CanLT}, and LCG~\citep{Gu2024LengthCG}, the key result is complementarity: adding LenVM to eight-call LCG raises the score from $71.3$ to $83.6$ and reduces average absolute deviation from $14.0\%$ to $5.8\%$. Appendix~\ref{app:additional_results} reports all two-, four-, and eight-call results.

\subsection{Application 2: Performance--Efficiency Trade-off}
\label{sec:tradeoff}

We next test whether LenVM supports \emph{continuous} control of the performance--efficiency trade-off without modifying the base generator.
Instead of enforcing a hard target length, we reweight the next-token distribution so that tokens predicted to lead to shorter continuations become more likely.

\paragraph{Value-guided exponential tilting.}
Let $p$ be the base model distribution over candidate set $\mathcal{V}_t$.
We define
\begin{equation}
  \min_{p'}\; \mathbb{E}_{p'}[\hat{v}(x)] - \frac{1}{\beta} D_{\mathrm{KL}}(p' \| p),
  \label{eq:kl_obj}
\end{equation}
whose solution is
\begin{equation}
  p'(x) = \frac{p(x)\exp(\beta\hat{v}(x))}
               {\sum_{x'\in\mathcal{V}_t} p(x')\exp(\beta\hat{v}(x'))}.
  \label{eq:tilted_dist}
\end{equation}
The derivation is provided in Appendix~\ref{app:exp_tilt_derivation}.
Here $\beta < 0$ controls the steering strength.
When $\beta = 0$, decoding reduces to the original model.
As $\beta$ becomes more negative, generation is increasingly biased toward shorter trajectories.

\begin{figure}[t]
    \centering
    \begin{subfigure}[b]{0.32\linewidth}
        \includegraphics[width=\linewidth]{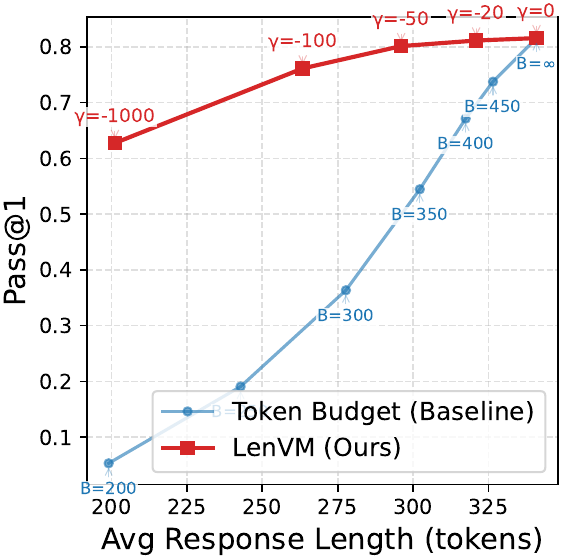}
        \caption{\scriptsize GSM8K; Base (Qwen2.5-3B-Instruct) \\+ LenVM (Qwen2.5-1.5B-Instruct)}
    \end{subfigure}
    \hfill
    \begin{subfigure}[b]{0.32\linewidth}
        \includegraphics[width=\linewidth]{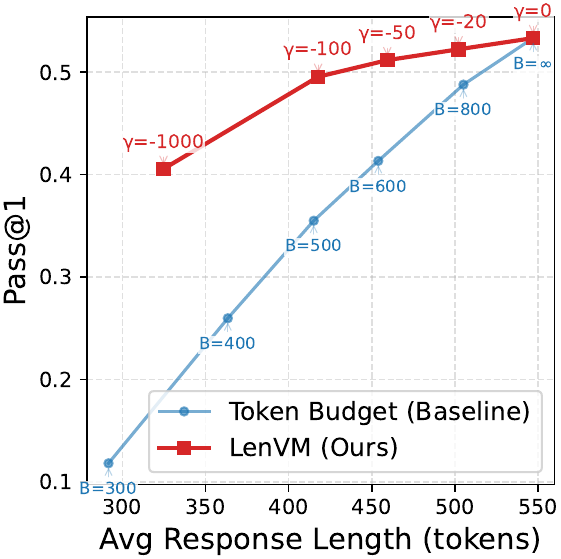}
        \caption{\scriptsize MATH500; Base (Qwen2.5-7B-Instruct) \\+ LenVM (Qwen2.5-1.5B-Instruct)}
    \end{subfigure}
    \hfill
    \begin{subfigure}[b]{0.32\linewidth}
        \includegraphics[width=\linewidth]{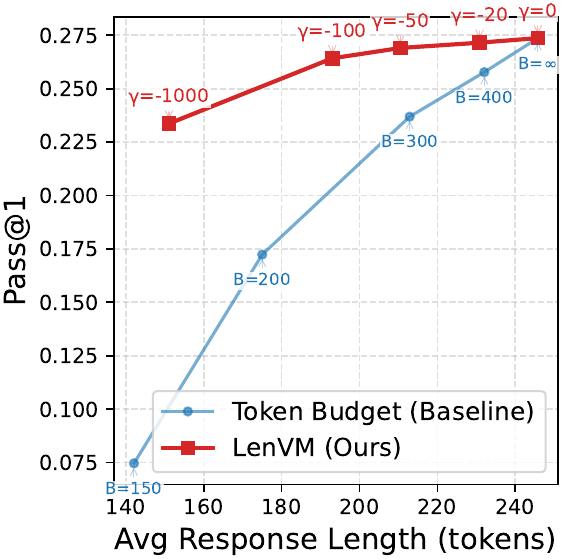}
        \caption{\scriptsize MathVista; Base (Qwen2.5-VL-7B-Instruct) + LenVM (Qwen2.5-VL-3B-Instruct)}
    \end{subfigure}
    \caption{\textbf{Performance (Pass@1) versus average generation length.} LenVM-guided exponential tilting (red) consistently outperforms a hard token-budget baseline (blue). The x-axis reports the average \emph{truncated} response length. Since the base generator is unchanged, the result shows that the model already has the ability to solve problems with shorter generations, and LenVM helps uncover those trajectories by reweighting token choices. Models in parentheses represent the initialization checkpoints.}
    \label{fig:length_performance_tradeoff}
\end{figure}

We evaluate on GSM8K~\citep{cobbe2021training}, MATH500~\citep{lightman2023let}, and MathVista~\citep{lu2023mathvista}.
As a baseline, we use a hard token budget that truncates generation once the token count exceeds threshold $B$ and marks the example as incorrect.
For fair comparison, the x-axis reports the average \emph{truncated} generation length for both methods.

Figure~\ref{fig:length_performance_tradeoff} shows that LenVM-guided decoding consistently achieves a better performance--efficiency frontier than hard truncation.
At the same average truncated length, LenVM retains much higher Pass@1.
For example, on GSM8K with Qwen2.5-3B-Instruct, at roughly 200 tokens the hard budget baseline achieves about 6\% Pass@1, whereas LenVM maintains about 63\%.
Because the base generator is never modified, this result indicates that the model itself already contains shorter successful reasoning paths.
LenVM simply biases decoding toward them.
Moreover, varying $\beta$ changes the frontier smoothly, confirming that LenVM provides a continuous knob for balancing performance and token budget.

\paragraph{Stronger matched-length baselines.}
We also compare with budget-aware prompting, which requests and enforces a token cap, and EOS calibration, which shifts the EOS logit. At matched average lengths, LenVM improves task accuracy on MATH500 and GSM8K, showing that its shorter outputs preserve answer quality. Full results and protocols appear in Appendices~\ref{app:additional_results} and~\ref{sec:add_exp_details}.

\paragraph{Comparison with a length-penalized RLVR model.}
On AIME2025, LenVM applied to Qwen3-30B-A3B-Instruct-2507 reaches a favorable matched-length point relative to ART, an RLVR model with a length penalty~\citep{Wu2026ArtEfficientReasoning}, and can further steer ART itself. This establishes compatibility rather than uniform superiority across training protocols; Appendix~\ref{app:additional_results} reports the results.

\paragraph{Policy dependence.}
LenVM is policy-conditioned. Our steering stays anchored to candidates from the base generator and evaluates nearby decoding shifts, not arbitrary cross-policy calibration; Appendix~\ref{sec:add_exp_details} discusses this scope.

\subsection{Application 3: Generation Length Prediction}
\label{sec:first_token_pred}

Beyond control, we test whether LenVM can predict generation length from the prompt boundary, i.e., from state $s_0$ before any response token is emitted.
This is useful for scheduling, batching, and memory planning at inference time.

To evaluate prediction accuracy, we use held-out prompts from math, code, and instruction-following data.
For each prompt, we sample $N=64$ completions with lengths $L_1,\dots,L_N$.
Because LenVM predicts discounted remaining horizon rather than raw token count, we evaluate in the same transformed space:
\begin{equation}
    u(L_i)=1-\gamma^{L_i}, \qquad
    \mu_u=\frac{1}{N}\sum_{i=1}^N u(L_i),
\end{equation}
and define the return-consistent ground-truth horizon as
\begin{equation}
    L_{\text{GT}}=\frac{\ln(1-\mu_u)}{\ln\gamma}.
\end{equation}
We then invert LenVM's prediction at $s_0$ into a length estimate $\hat{L}$ and report Mean Relative Error (MRE):
\begin{equation}
\mathrm{MRE}(\hat L, L_{\text{GT}})
\triangleq
\mathbb{E}\left[\frac{|\hat L-L_{\text{GT}}|}{L_{\text{GT}}}\right].
\label{eq:mre}
\end{equation}
Further discussion of this target appears in Appendix~\ref{app:length_underestimation}.

\paragraph{Specialized prediction baselines.}
We compare with EGTP, which predicts total length from entropy-weighted prompt states, and PLP, which extends this predictor to generated prefixes~\citep{Xie2026PredictingLO}. Appendix~\ref{sec:add_exp_details} gives the matched training and evaluation protocol.

\begin{table}[!t]
  \centering
  \caption{%
    \textbf{Generation length prediction.}
    Panel (a) reports prompt-boundary Mean Relative Error (MRE) across scales and domains; IF is instruction following. Panel (b) compares specialized predictors. MAE is Mean Absolute Error, $\rho$ is Spearman correlation, and W128 is Within-128 accuracy.
  }
  \label{tab:first_token_mre}
  \label{tab:prediction_baselines}
  \begin{minipage}[t]{0.40\textwidth}
  \centering
  \textbf{(a) Scaling across domains}\\[2pt]
  \small
  {\setlength{\tabcolsep}{5pt}
  \begin{tabular}{crrr}
    \toprule
    \multirow{2}{*}{Model size} & \multicolumn{3}{c}{MRE $\downarrow$} \\
    \cmidrule(lr){2-4}
    & Math & Code & IF \\
    \midrule
    1.5B  & 17.0\% & 29.0\% & 33.0\% \\
    3B    & 13.6\% & 24.0\% & 27.2\% \\
    7B    & 11.0\% & 19.5\% & 23.0\% \\
    14B   & 10.4\% & 17.0\% & 19.8\% \\
    32B   & 9.8\%  & 14.9\% & 17.1\% \\
    \bottomrule
  \end{tabular}
  }
  \end{minipage}
  \hfill
  \begin{minipage}[t]{0.57\textwidth}
  \centering
  \textbf{(b) Specialized baselines}\\[2pt]
  \footnotesize
  {\setlength{\tabcolsep}{4.8pt}
  \begin{tabular}{llrrrr}
    \toprule
    Method & Mode & MRE $\downarrow$ & MAE $\downarrow$ & $\rho$ $\uparrow$ & W128 $\uparrow$ \\
    \midrule
    EGTP & Prompt & 32.56 & 228.31 & 0.344 & 35.34 \\
    LenVM & Prompt & \textbf{26.90} & \textbf{194.39} & \textbf{0.575} & \textbf{42.65} \\
    PLP & Prefix & 26.71 & 218.67 & 0.746 & 30.74 \\
    LenVM & Prefix & \textbf{15.10} & \textbf{155.56} & \textbf{0.783} & \textbf{60.70} \\
    \bottomrule
  \end{tabular}
  }
  \end{minipage}
\end{table}

As shown in Table~\ref{tab:first_token_mre}(a), LenVM predicts generation horizon from the prompt boundary with low relative error across all domains, and accuracy improves consistently with scale.
At 32B, MRE reaches 9.8\% on math, 14.9\% on code, and 17.1\% on instruction following.
This shows that LenVM is not only useful for online control, but also captures predictive information about expected generation length before decoding begins.

\paragraph{Comparison with specialized predictors.}
Table~\ref{tab:first_token_mre}(b) reports results using a frozen Qwen2.5-7B-Instruct generator. Prompt metrics cover $8192$ held-out DeepMath-103K prompts, while progressive metrics cover approximately $108{,}000$ prefix states. LenVM improves all four metrics in both settings, reducing prompt MRE from $32.56$ to $26.90$ and progressive MRE from $26.71$ to $15.10$.

\subsection{Scalability of LenVM}
\label{sec:scalability}

A key advantage of LenVM is that its supervision is annotation-free, dense, and easy to scale.
We therefore study scaling along three axes:
\textbf{(i) model size},
\textbf{(ii) number of training questions}, and
\textbf{(iii) number of samples per question}.

\begin{figure*}[t]
  \centering
  \begin{subfigure}[b]{0.325\linewidth}
    \includegraphics[width=\linewidth]{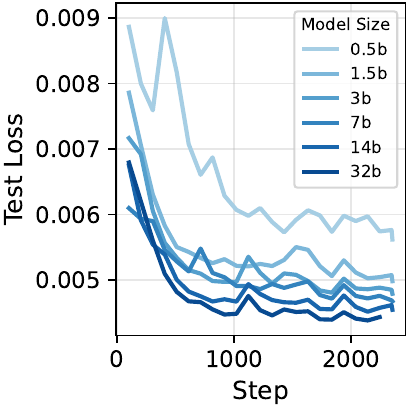}
    \caption{Model size}
    \label{fig:scale_model_size}
  \end{subfigure}
  \hfill
  \begin{subfigure}[b]{0.33\linewidth}
    \includegraphics[width=\linewidth]{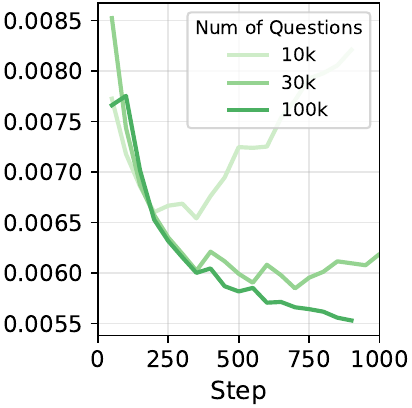}
    \caption{Number of training questions}
    \label{fig:scale_num_questions}
  \end{subfigure}
  \hfill
  \begin{subfigure}[b]{0.33\linewidth}
    \includegraphics[width=\linewidth]{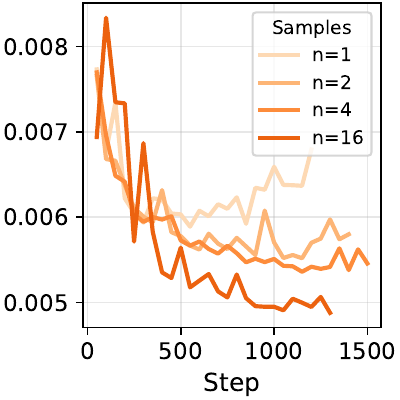}
    \caption{Samples per question}
    \label{fig:scale_samples}
  \end{subfigure}
  \caption{\textbf{LenVM scales along three axes.} Validation loss decreases consistently as we increase (\textbf{a}) model size, (\textbf{b}) number of training questions, and (\textbf{c}) samples per question.}
  \label{fig:scalability}
\end{figure*}

Figure~\ref{fig:scalability} shows consistent improvements along all three axes.
Larger models achieve lower validation loss, matching the gains in prompt-boundary prediction accuracy from Table~\ref{tab:first_token_mre}.
Increasing the number of training prompts also improves performance, indicating that LenVM benefits from broader coverage of tasks and trajectory types.
Finally, increasing the number of sampled completions per prompt further reduces validation loss, showing that supervision can be scaled not only through more prompts but also through more sampled trajectories from the same prompts.
Overall, these results support LenVM as a scalable value-pretraining objective.

\subsection{Inference Cost}
\label{sec:inference_cost}

We score the generator's top-$K$ candidates in one packed SGLang call~\citep{Zheng2024SGLang}, cache LenVM key-value states, and skip value scoring when generator entropy is below $0.5$. Table~\ref{tab:latency} shows that skipping activates LenVM on $19.5\%$ of decoding steps and reduces measured end-to-end latency from $4.55\times$ to $1.90\times$. Appendix~\ref{app:additional_results} gives the full protocol and candidate-scoring times.

\begin{table}[H]
  \centering
  \caption{\textbf{End-to-end decoding latency.} Active steps are decoding steps on which LenVM is evaluated.}
  \label{tab:latency}
  \begin{tabular}{lrr}
    \toprule
    Setting & Active steps & Latency \\
    \midrule
    Vanilla decoding & $0\%$ & $1.00\times$ \\
    LenVM at every step & $100\%$ & $4.55\times$ \\
    LenVM with entropy skip & $19.5\%$ & $1.90\times$ \\
    \bottomrule
  \end{tabular}
\end{table}

\paragraph{Additional analyses.}
Appendix~\ref{sec:lvm_ablation} studies target representation, batching, discount factor, numerical precision, and candidate-set size. Section~\ref{sec:deep_thinking_token} analyzes tokens associated with shifts toward shorter or longer predicted horizons.

\paragraph{Scope and limitations.}
Guided decoding remains $1.9\times$ slower than vanilla decoding in our setup, with hardware and serving-load sensitivity. LenVM is policy-conditioned, so calibration under unrelated generators remains open. LIFEBench-token uses token constraints and is not directly comparable with the original leaderboard. Empirical RL fine-tuning with LenVM is future work.

\section{Qualitative Case Study: Length Tokens as Markers of Length Shifts}
\label{sec:deep_thinking_token}

\begin{figure}[!t]
  \centering
  \begin{subfigure}{0.49\linewidth}
    \centering
    \includegraphics[width=0.92\linewidth]{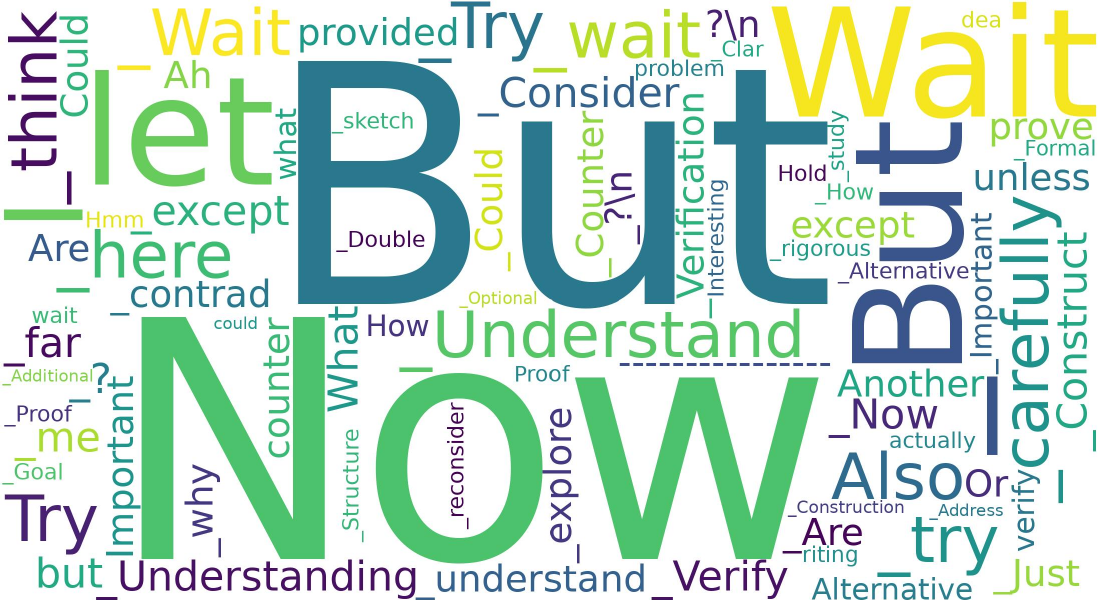}
    \caption{Positive length tokens: longer-horizon shifts}
    \label{fig:positive_length_token}
  \end{subfigure}
  \hfill
  \begin{subfigure}{0.49\linewidth}
    \centering
    \includegraphics[width=0.92\linewidth]{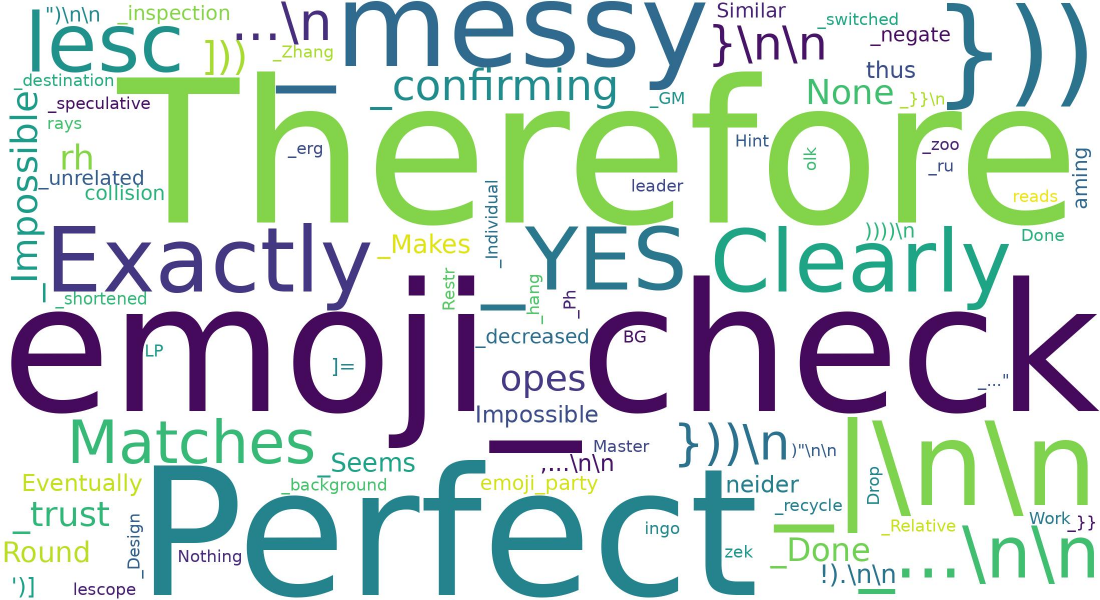}
    \caption{Negative length tokens: shorter-horizon shifts}
    \label{fig:negative_length_token}
  \end{subfigure}
  \caption{\textbf{Length tokens as markers of horizon shifts.} Frequent tokens with TD residual $s_t$ above $0.01$ or below $-0.01$ on 8k DeepMath-103K examples. Token size is proportional to frequency.}
  \label{fig:token_length_jump_wordcloud}
\end{figure}

We call tokens associated with changes in LenVM's predicted remaining horizon \emph{length tokens}. At step $t$, their temporal-difference residual is
\begin{equation}
s_t \triangleq r_{t-1} + \gamma V_t - V_{t-1},
\qquad r_{t-1}=-(1-\gamma),
\label{eq:value_td_score}
\end{equation}
where $V_t \triangleq V_\theta(s_t)$ and $\gamma$ is the training discount factor. Positive and negative $s_t$ mark longer and shorter horizons, respectively. Figure~\ref{fig:positive_length_token} contrasts reasoning pivots such as \texttt{wait}, \texttt{think}, and \texttt{try} with closure tokens such as \texttt{therefore}, \texttt{clearly}, and \texttt{perfect}.

\section{Ablations}
\label{sec:lvm_ablation}

We ablate five design choices in LenVM:
\textbf{(i) length-space representation},
\textbf{(ii) batch construction strategy},
\textbf{(iii) discount factor $\gamma$}, and
\textbf{(iv) numerical precision} (\textsc{fp16}/\textsc{bf16}/\textsc{fp32}), and
\textbf{(v) candidate-set size $K$}.
Unless otherwise specified, all experiments keep the model, optimizer, and evaluation protocol fixed.

\begin{figure}[!t]
    \centering
    \begin{subfigure}{0.33\textwidth}
        \includegraphics[width=\linewidth]{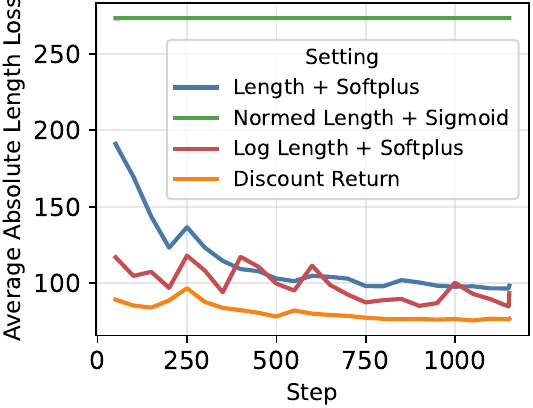}
        \caption{Length-space representation}
        \label{fig:length_space_ablation}
    \end{subfigure}\hfill
    \begin{subfigure}{0.33\textwidth}
        \includegraphics[width=\linewidth]{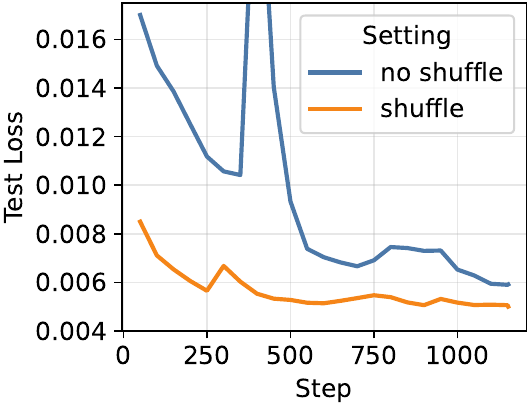}
        \caption{Shuffling vs.\ grouped batching}
        \label{fig:shuffle_ablation}
    \end{subfigure}\hfill
    \begin{subfigure}{0.33\textwidth}
        \includegraphics[width=\linewidth]{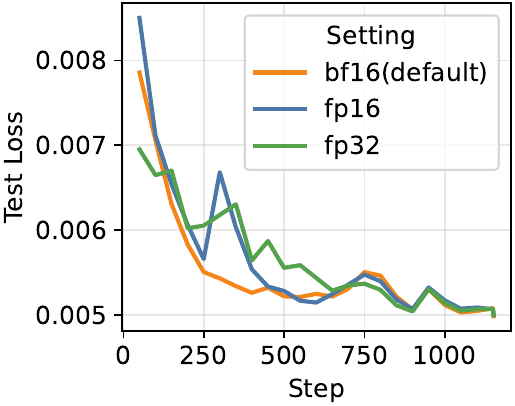}
        \caption{Numerical precision}
        \label{fig:precision_ablation}
    \end{subfigure}
    \caption{\textbf{Ablations on LenVM design choices.} Comparison of (a) target representation, (b) batch construction strategy, and (c) numerical precision.}
    \label{fig:combined_ablation}
\end{figure}

\begin{figure}[!t]
    \centering
    \begin{subfigure}{0.24\textwidth}
        \includegraphics[width=\linewidth]{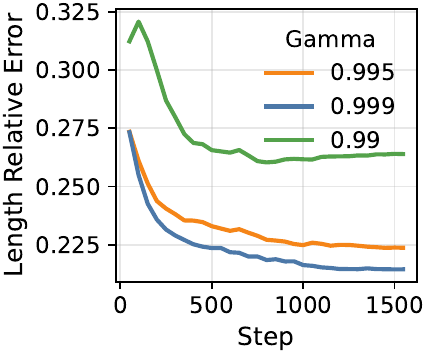}
        \caption{0\%}
    \end{subfigure}
    \begin{subfigure}{0.24\textwidth}
        \includegraphics[width=\linewidth]{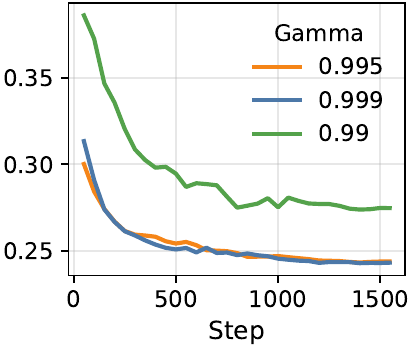}
        \caption{25\%}
    \end{subfigure}
    \begin{subfigure}{0.24\textwidth}
        \includegraphics[width=\linewidth]{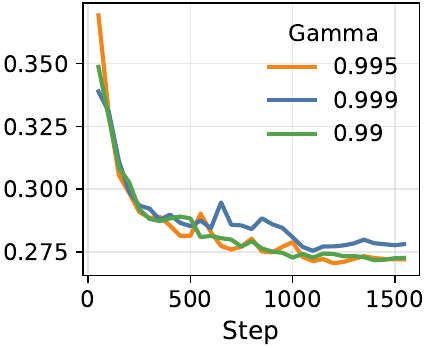}
        \caption{50\%}
    \end{subfigure}
    \begin{subfigure}{0.24\textwidth}
        \includegraphics[width=\linewidth]{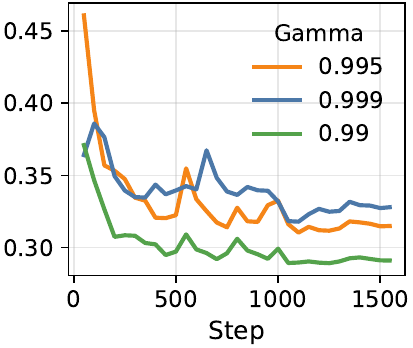}
        \caption{75\%}
    \end{subfigure}
    \caption{\textbf{Ablation on the discount factor $\gamma$.} Prediction error at different relative decoding positions. Larger $\gamma$ performs better earlier in generation, while smaller $\gamma$ performs better closer to termination.}
    \label{fig:gamma_ablation}
\end{figure}

\subsection{Length-Space Representation}
We first compare four target parameterizations in Figure~\ref{fig:length_space_ablation}:
\textbf{(1) Length + Softplus},
\textbf{(2) Normalized Length + Sigmoid},
\textbf{(3) Log Length + Softplus}, and
our proposed \textbf{(4) Discount Return + Sigmoid}.
Among them, \textbf{Discount Return + Sigmoid} consistently achieves the lowest average absolute length error throughout training. Raw-length and normalized-length regression perform worse, partly because generation lengths are highly long-tailed.
In particular, normalizing by the maximum length compresses the dense short-length region into a narrow interval near zero, where discrimination becomes difficult.
Log-length regression is a stronger baseline and substantially improves over raw or normalized length, but still underperforms discount return.
Unlike a static scale transform such as log length, discount return is aligned with the stepwise Bellman recursion of autoregressive decoding, yielding a more coherent token-level training signal. This result shows that for token-level length modeling, directly regressing raw length or normalized length is ineffective, while \textit{the discounted-return formulation provides a better-conditioned and more effective target.}

\subsection{Batch Construction Strategy}
We compare fully shuffled training against grouped batching that keeps multiple completions from the same prompt together, while holding the total number of data per update fixed.
As shown in Figure~\ref{fig:shuffle_ablation}, shuffling consistently improves evaluation loss.
This suggests that, unlike the reward model, grouping multiple samples from the same prompt within a batch is unnecessary and can mildly hurt generalization.

\subsection{Discount Factor \texorpdfstring{$\gamma$}{gamma}}
\label{sec:gamma_ablation}
The discount factor $\gamma$ determines how remaining horizon is distributed in the target space, and therefore where prediction resolution is concentrated along the trajectory.
Larger $\gamma$ compresses long horizons more aggressively, which tends to improve prediction earlier in generation.
Smaller $\gamma$ allocates more resolution near termination, improving prediction later in the trajectory.
To evaluate this trade-off, we measure prediction error at relative decoding positions 0\%, 25\%, 50\%, and 75\%.
Figure~\ref{fig:gamma_ablation} shows the expected pattern: larger $\gamma$ performs better at earlier positions, while smaller $\gamma$ performs better closer to the end.
Thus, $\gamma$ acts as a simple knob for balancing early-horizon compression against late-horizon discrimination. In practice, we choose an intermediate $\gamma$ to balance early-horizon compression against late-horizon discrimination.
Specifically, we set $\gamma$ such that the 99th-percentile generation length $L_{0.99}$ satisfies $1-\gamma^{L_{0.99}} = 0.99$, so that nearly all observed horizons are mapped into the high-resolution region of the target space.

\subsection{Numerical Precision}
We compare \textsc{fp16}, \textsc{bf16}, and \textsc{fp32} under the same training and evaluation setup.
Figure~\ref{fig:precision_ablation} shows nearly identical loss curves across all three formats, with no substantial difference in convergence or final performance.
This indicates that LenVM is numerically stable under common floating-point precisions.
Appendix~\ref{app:precision_analysis_details} provides further analysis of finite-precision effects.

\subsection{Candidate-Set Size}
\label{sec:candidate_size_ablation}

Candidate-set size affects hard target matching and soft efficiency steering differently. Increasing $K$ from $1$ to $128$ raises the LIFEBench-token score from $29.3$ to $72.1$ and reduces deviation from $100\%$ to $19\%$. On MATH500 with $\beta=-100$, most of the change occurs between $K=1$ and $K=2$; beyond $K=2$, average length and Pass@1 remain stable. Appendix~\ref{app:additional_results} reports the complete ablation and packed forward times.

\section{Conclusion}
We proposed \textbf{LenVM}, a token-level value model for remaining generation length.
By assigning a constant per-token reward and predicting the resulting discounted return, LenVM places length modeling in a standard value-learning framework with bounded, dense, and annotation-free supervision.
Our experiments show that this simple formulation is both useful and scalable.
LenVM supports precise length-controlled generation, exposes a smooth performance--efficiency trade-off without modifying the base generator, predicts expected generation horizon, and improves consistently with model scale and automatically collected supervision.
Additional experiments show that LenVM complements iterative controllers, matched-length baselines, and specialized predictors, while entropy skipping reduces overhead. It also steers a length-penalized RLVR model, and the case study identifies token-level markers of horizon shifts.
These results suggest that discounted length value is an effective target for token-level length modeling and a promising substrate for scalable value pretraining.


\bibliography{main}

\clearpage
\appendix
\section*{Appendix}
\addcontentsline{toc}{section}{%
  {\large Appendix}%
}
\section{LIFEBench-token Evaluation Details}
\label{app:lifebench_details}

\paragraph{Benchmark overview.}
LIFEBench \citep{zhang2025lifebench} is a comprehensive benchmark designed to evaluate the ability of large language models to follow explicit length instructions across diverse tasks and length scales.
It comprises $10,800$ ($360$ data, $3$ length constraints, $10$ length target) instances spanning four task categories:
\begin{itemize}
    \item \textbf{Question Answering (QA):} Open-ended questions drawn from Quora, Stack Exchange, WikiHow, and Yahoo Answers (English), and from Zhihu and WikiHow (Chinese).
    \item \textbf{Summarization:} Long documents (10,000--15,000 words) sourced from scientific papers, government reports, book chapters, and Wikipedia (English), and from corporate annual reports, Wikipedia, and XueXiQiangGuo (Chinese).
    \item \textbf{Reasoning:} Eighty problems (40 per language) generated via GPT-4o, covering six reasoning types: deductive, inductive, abductive, analogical, causal, and probabilistic.
    \item \textbf{Creative Generation:} Writing prompts derived from movie scripts, story corpora, and news datasets (English), and from web novels, social commentary, and WeChat public articles (Chinese).
\end{itemize}

\paragraph{Length constraints.}
The original benchmark focuses on word and character lengths. Each instance is paired with one of three constraint types (\emph{Equal To}, \emph{At Most}, and \emph{At Least}) and one of ten target lengths: 16, 32, 64, 128, 256, 512, 1,024, 2,048, 4,096, and 8,192 words for English or characters for Chinese. In LIFEBench-token, we preserve the tasks, target grid, and metrics but apply each constraint to the model-specific token count. Specifically, we prompt the model to generate a response whose total token count is equal to, at most, or at least the target. Tokens directly determine autoregressive decoding steps and key-value cache growth, but this change of unit means that our absolute scores are not directly comparable with the original word-level or character-level leaderboard.

\paragraph{Evaluation metrics.}
Two metrics are used:

\begin{itemize}
    \item \textbf{Length Deviation (LD):} For example $i$,
    \begin{equation}
        \mathrm{LD}_i = \frac{L_{\text{output},i} - L_{\text{constraint},i}}{L_{\text{constraint},i}},
        \label{eq:lifebench_ld}
    \end{equation}
    where $L_{\text{output},i}$ is the actual output length and $L_{\text{constraint},i}$ is the target length for example $i$.
    Positive values indicate over-generation; negative values indicate under-generation.
    Following the official aggregation protocol for the \emph{Equal To} setting, we report the mean of $|\mathrm{LD}|$ over evaluated examples. The reported LD is therefore nonnegative and equals the mean absolute relative deviation.

    \item \textbf{Length Score (LS):} A bounded score in $[0, 100]$ derived from LD via an exponential penalty function that is adapted to each constraint type:
    \begin{equation}
        \mathrm{LS}_i =
        \begin{cases}
        100 \cdot e^{k_1 \mathrm{LD}_i}, & \mathrm{LD}_i < 0 \quad (\text{Equal To / At Least}), \\
        100 \cdot e^{-k_2 \mathrm{LD}_i}, & \mathrm{LD}_i > 0 \quad (\text{Equal To / At Most}), \\
        100, & \text{otherwise},
        \end{cases}
        \label{eq:lifebench_ls}
    \end{equation}
    with $k_1 = 5$ and $k_2 = 2$. The reported score over $N$ examples is
    \begin{equation}
        \overline{\mathrm{LS}} = \frac{1}{N}\sum_{i=1}^{N}\mathrm{LS}_i.
        \label{eq:lifebench_ls_aggregate}
    \end{equation}
    A score of 100 indicates perfect adherence; lower scores reflect larger deviations.
\end{itemize}

\paragraph{Implementation details.}
Each prompt follows the template \texttt{[Instruction] + [Requirement]}, where the instruction specifies the task and input content, and the requirement states the length constraint in natural language (e.g., ``Your answer must be equal to 512 tokens.'').

\section{Additional Experimental Details}
\label{sec:add_exp_details}

We train LenVM using LlamaFactory.
The discount factor $\gamma$ is set to $0.997$ for \textbf{Qwen2.5-Instruct} and \textbf{Qwen2.5-VL-Instruct}, and to $0.9998$ for \textbf{Qwen3-Instruct}.
We set $\lambda=1$ for GAE.

For the ablation and scalability experiments, we sample 100k examples from each dataset, except for math, where we use all 95k available examples.
For the Length-Controlled Generation and Performance--Efficiency Trade-off experiments, we use the full training sets of all three datasets.
We randomly sample 8k examples for validation during training.
Unless otherwise specified, data sampling uses temperature $1.0$ and top-p $1.0$, and we sample up to 16 completions per prompt.

We train LenVM for 2 epochs with learning rate $2\times 10^{-5}$, batch size $1024$, and BF16 precision.

In \textbf{Length-Controlled Generation}, we use temperature $1.0$, top-p $0.999$, and top-k $15$.
All baseline settings follow the original LIFEBench paper, except that constraints are applied to token count in our LIFEBench-token experiments.
We use the official tokenizer of each model, or the token count returned by the corresponding API response, to measure generation length.
Although tokenization schemes differ across models, all methods are evaluated under the same token-based constraint setting.

\paragraph{Exact-control baseline implementations.}
All exact-control experiments in Table~\ref{tab:lifebench_length_control} use Qwen2.5-7B-Instruct, the same LIFEBench-token prompts and targets, and the Qwen2.5 tokenizer for both feedback and evaluation. Unless a method specifies its own decoding rule, generation uses temperature $1.0$, top-p $0.999$, and top-k $15$. For Prompt Best-of-$N$, we independently sample $N\in\{2,4,8\}$ responses from the ordinary target-length prompt and select the response minimizing $|L(y)-L^\star|$, where $L^\star$ is the target token count. For CAPEL~\citep{Xie2025PromptBasedOE}, we adapt the published one-pass prompt by requesting token control and appending its countdown markers and explicit counting rules. We remove the countdown markup before computing task metrics and measure the remaining response with the Qwen2.5 tokenizer.

For Dynamic Feedback~\citep{Xiao2026CanLT}, we begin with the target-length prompt, pause at detected sentence boundaries, compute the number of generated tokens, and append a feedback string of the form \texttt{<used\_tokens=$n$>} before resuming. The prompt explains that this value is supplied by an external counter and should be used to meet the target. Each resume is a new generator call, and the two-, four-, and eight-call settings cap the total number of generation segments accordingly. For LCG~\citep{Gu2024LengthCG}, the first call produces a length-conditioned response. Each subsequent call receives the previous response together with its signed token-count error and is instructed to add or delete the required amount. We retain the method's importance-guided proposal and Metropolis-Hastings acceptance rule, which favors candidates with better target-length fit while preserving the generator's response-quality score. The reported call count includes the initial response.

\paragraph{Combining LCG with LenVM.}
LCG with LenVM keeps the same outer proposal, scoring, and acceptance procedure as LCG. The only change is that every initial-generation or refinement call uses the Equal-To LenVM rule from Section~\ref{sec:length_controlled_generation} when decoding the proposed response. Thus the outer controller still revises complete responses using observed length error, while LenVM guides token selection within each proposal toward the remaining target horizon. We use the same call budgets and $K=15$ candidate set for the paired LCG and LCG with LenVM comparisons.

In the \textbf{Performance--Efficiency Trade-off} experiments, we use temperature $1.0$ and top-p $1.0$.
For LenVM guided decoding, we set min-p to $0.01$ to reduce the candidate set size and avoid excessively large LenVM forward batches.
We sample 64 completions for each question.

\paragraph{Matched-length baseline implementations.}
Budget-aware prompting prepends the instruction \texttt{Solve the problem within $B$ tokens} and sets \texttt{max\_tokens}$=B$ so that the requested and enforced budgets agree. We sweep $B$ to obtain multiple accuracy and average-length points. For EOS calibration, if $z_{\mathrm{EOS}}$ is the generator's EOS logit, we decode with $z'_{\mathrm{EOS}}=z_{\mathrm{EOS}}+b$ and leave all other logits unchanged. Sweeping $b$ changes the stopping tendency without changing model parameters. For each baseline point, we report the measured LenVM point with the smallest absolute difference in average output length. All methods use Qwen2.5-7B-Instruct, identical benchmark prompts and answer extraction, temperature $1.0$, and top-p $1.0$.

For the RLVR comparison, we evaluate the released ART checkpoint based on Qwen3-30B-A3B-Instruct-2507~\citep{Wu2026ArtEfficientReasoning} on AIME2025 with the same prompts, sampling configuration, and answer checker used for its unmodified base model. Base with LenVM changes only the decoding distribution through $\beta$. ART with LenVM applies the same exponential tilting to ART without further parameter updates, so the points isolate decoding-time steering on top of the length-penalized policy.

\paragraph{Policy dependence.}
LenVM estimates $V^\pi(s_t)=\mathbb{E}_\pi[G_t\mid s_t]$ for the rollout policy used to construct its supervision, so we do not assume calibration under arbitrary unrelated generators or large policy shifts. In our guided decoding experiments, candidates come from the base generator and exponential tilting remains anchored to its token distribution. This limits, but does not eliminate, off-policy mismatch, and the reported curves test nearby decoding-policy shifts only.

\paragraph{Length-prediction baseline implementations.}
We implement EGTP and PLP following \citet{Xie2026PredictingLO} with Qwen2.5-7B-Instruct frozen. EGTP computes next-token entropy at every prompt position, applies softmax-normalized entropy weights to the corresponding hidden states, and sends the pooled representation to a lightweight length-prediction head. PLP uses the same prompt representation together with hidden states from the generated prefix and predicts the number of remaining tokens at each sampled prefix. The prediction heads are trained on DeepMath-103K training trajectories without updating the generator. We evaluate EGTP and prompt-boundary LenVM on the same $8192$ held-out prompts. For progressive prediction, we sample prefix states from their held-out completions, producing approximately $108{,}000$ matched states for PLP and LenVM. All methods use the same realized completion lengths as targets and the same MRE, MAE, Spearman correlation, and Within-128 calculations.

For the latency study, the generator is \textbf{Qwen2.5-7B-Instruct} and LenVM is initialized from \textbf{Qwen2.5-1.5B-Instruct}. Candidate next states are evaluated in one packed SGLang call~\citep{Zheng2024SGLang}, and LenVM key-value states are cached across decoding steps. The entropy-skipping variant invokes LenVM only when the generator entropy is at least $0.5$.

\section{Additional Experimental Results}
\label{app:additional_results}

\subsection{End-to-End Latency}

For the latency study, we use Qwen2.5-7B-Instruct as the generator and Qwen2.5-1.5B as LenVM. Candidate tokens are drawn from the generator's top-$K$ set, so LenVM does not expand the candidate pool. They are scored in one packed SGLang call~\citep{Zheng2024SGLang}, each candidate attends to the shared prefix but not to other candidates, and LenVM key-value states are cached. We skip LenVM scoring when generator entropy is below $0.5$ because low-entropy steps offer little room for length guidance to change the selected token. Table~\ref{tab:latency} reports the resulting end-to-end latency. This remains a meaningful serving overhead.

\subsection{Matched-Length Quality Baselines}

Budget-aware prompting requests a solution within $B$ tokens and sets \texttt{max\_tokens}$=B$. EOS calibration instead adjusts the EOS probability to move along the accuracy and length curve. All methods in Table~\ref{tab:matched_length_quality} use Qwen2.5-7B-Instruct and are compared at their nearest-length operating points. LenVM improves over both baselines on MATH500 and over budget-aware prompting on GSM8K.

\begin{table}[H]
  \centering
  \caption{\textbf{Task accuracy at matched average output lengths.}}
  \label{tab:matched_length_quality}
  \begin{tabular}{llrr}
    \toprule
    Dataset & Method & Avg. tokens & Pass@1 $\uparrow$ \\
    \midrule
    MATH500 & Budget prompt & 333 & 39.66 \\
    MATH500 & LenVM & 317 & \textbf{47.50} \\
    MATH500 & Budget prompt & 412 & 48.63 \\
    MATH500 & LenVM & 397 & \textbf{50.78} \\
    MATH500 & EOS calibration & 307 & 28.56 \\
    MATH500 & LenVM & 306 & \textbf{45.59} \\
    \midrule
    GSM8K & Budget prompt & 198 & 81.00 \\
    GSM8K & LenVM & 192 & \textbf{86.28} \\
    GSM8K & Budget prompt & 211 & 86.88 \\
    GSM8K & LenVM & 207 & \textbf{87.88} \\
    \bottomrule
  \end{tabular}
\end{table}

\subsection{Full LIFEBench-token Comparison}

Table~\ref{tab:lifebench_full_control} reports all tested inference budgets for prompt Best-of-$N$, CAPEL~\citep{Xie2025PromptBasedOE}, Dynamic Feedback~\citep{Xiao2026CanLT}, LCG~\citep{Gu2024LengthCG}, and LCG with LenVM. A round is one generation or refinement call. The methods therefore do not have equal inference cost. Across two, four, and eight rounds, inserting LenVM into LCG consistently lowers average absolute deviation and raises the length score.

\begin{table}[H]
  \centering
  \caption{\textbf{Full LIFEBench-token comparison across inference budgets.} LD is Length Deviation and is reported as the mean absolute relative deviation. Length Score is a bounded adherence score in $[0,100]$, with 100 indicating perfect adherence. Lower LD and higher Length Score are better.}
  \label{tab:lifebench_full_control}
  \begin{tabular}{llrr}
    \toprule
    Method & Protocol & LD $\downarrow$ & Score $\uparrow$ \\
    \midrule
    LCG + LenVM & 8 rounds & \textbf{5.8\%} & \textbf{83.6} \\
    LCG + LenVM & 4 rounds & 8.6\% & 80.5 \\
    LCG + LenVM & 2 rounds & 10.7\% & 73.8 \\
    LenVM & 1 round & 44.0\% & 64.8 \\
    \midrule
    LCG & 8 rounds & 14.0\% & 71.3 \\
    LCG & 4 rounds & 15.3\% & 69.0 \\
    LCG & 2 rounds & 20.3\% & 60.6 \\
    Dynamic Feedback & 8 rounds & 15.9\% & 66.6 \\
    Dynamic Feedback & 4 rounds & 25.3\% & 46.5 \\
    Dynamic Feedback & 2 rounds & 42.9\% & 30.4 \\
    Prompt & Best-of-8 & 28.7\% & 58.1 \\
    Prompt & Best-of-4 & 38.4\% & 47.2 \\
    Prompt & Best-of-2 & 50.2\% & 38.4 \\
    CAPEL & 1 round & 140.8\% & 28.8 \\
    \bottomrule
  \end{tabular}
\end{table}

\subsection{RLVR Length-Penalty Comparison}

We compare LenVM with ART, which trains Qwen3-30B-A3B-Instruct-2507 using RLVR and a length penalty~\citep{Wu2026ArtEfficientReasoning}. Table~\ref{tab:art_comparison} separates two questions. The first group compares nearby measured operating points from different protocols. Base + LenVM reaches $54.17$ Pass@1 at $5611$ tokens, compared with $51.25$ at $5599$ tokens for ART. The second group applies LenVM directly to ART. Stronger steering shortens the average response from $5599$ to $4949$ tokens while accuracy decreases from $51.25$ to $46.25$, exposing a continuous quality and length trade-off.

\begin{table}[H]
  \centering
  \caption{\textbf{Comparison with RLVR plus a length penalty on AIME2025.} RLVR denotes reinforcement learning with verifiable rewards, $\beta$ is the LenVM steering coefficient.}
  \label{tab:art_comparison}
  \begin{tabular}{lrrr}
    \toprule
    Method & $\beta$ & Avg. tokens & Pass@1 $\uparrow$ \\
    \midrule
    Base & 0 & 6595 & 60.00 \\
    ART & 0 & 5599 & 51.25 \\
    Base + LenVM & $-200$ & 5611 & \textbf{54.17} \\
    Base + LenVM & $-300$ & \textbf{5422} & \textbf{52.92} \\
    \midrule
    ART & 0 & 5599 & 51.25 \\
    ART + LenVM & $-10$ & 5391 & 49.58 \\
    ART + LenVM & $-25$ & 5255 & 46.67 \\
    ART + LenVM & $-50$ & 4949 & 46.25 \\
    \bottomrule
  \end{tabular}
\end{table}

\subsection{Candidate-Set Size}

Under hard constraints, a larger candidate pool increases the chance that a next state matches the target horizon. Under soft steering, exponential tilting remains anchored to the generator distribution, so low-probability candidates remain suppressed. Table~\ref{tab:candidate_size} shows that increasing $K$ from $1$ to $128$ raises the LIFEBench-token score from $29.3$ to $72.1$ and reduces deviation from $100.2\%$ to $18.5\%$, although intermediate values are not strictly monotone. On MATH500, most of the change occurs between $K=1$ and $K=2$.

\begin{table}[H]
  \centering
  \caption{\textbf{Sensitivity to candidate-set size.} $K$ is the candidate-set size (top-k), LB-token denotes LIFEBench-token, and its score is the Length Score in $[0,100]$, where higher is better. $\beta$ is the LenVM steering coefficient. MATH500 uses $\beta=-100$.}
  \label{tab:candidate_size}
  \begin{adjustbox}{max width=\textwidth}
  \begin{tabular}{rrrrr}
    \toprule
    $K$ & LB-token deviation $\downarrow$ & LB-token score $\uparrow$ & MATH500 Pass@1 $\uparrow$ & MATH500 avg. tokens \\
    \midrule
    1 & 100.2\% & 29.3 & 55.84 & 636.95 \\
    2 & 30.6\% & 56.5 & 52.78 & 435.56 \\
    8 & 38.6\% & 56.1 & 50.72 & 395.08 \\
    15 & 26.5\% & 63.5 & 50.72 & 391.60 \\
    32 & 21.4\% & 69.7 & 50.03 & 392.26 \\
    128 & \textbf{18.5\%} & \textbf{72.1} & 50.88 & 393.76 \\
    \bottomrule
  \end{tabular}
  \end{adjustbox}
\end{table}

\subsection{Packed Candidate-Scoring Time}

All candidate next states are scored in one packed SGLang call~\citep{Zheng2024SGLang}, with a tree-structured attention mask that lets each candidate attend to the shared prefix but not to other candidates. Table~\ref{tab:packed_forward_time} reports LenVM forward time per active decoding step. Within the tested range, time stays between $32.18$ and $34.16$ ms as $K$ increases from $2$ to $5$. This rules out a naive sequence of $K$ independent LenVM calls in this implementation, but it does not imply that runtime is independent of $K$ outside the tested range.

\begin{table}[H]
  \centering
  \caption{\textbf{LenVM packed forward time per active decoding step.}}
  \label{tab:packed_forward_time}
  \begin{tabular}{rr}
    \toprule
    Candidate-set size $K$ & LenVM forward time \\
    \midrule
    2 & 32.18 ms \\
    3 & 34.16 ms \\
    5 & 33.49 ms \\
    \bottomrule
  \end{tabular}
\end{table}

\section{LenVM as a Length-Specific Value Function in RL}
\label{sec:rl_reward_signal}

This section makes explicit the RL interpretation of LenVM and clarifies two distinct ways in which it can interact with reinforcement learning.
First, the return constructed in Section~\ref{sec:length_value_model} is already a valid discounted reward process, so LenVM can be viewed as a value function for a token-level length objective.
Second, because LenVM defines a state-dependent notion of progress toward termination, it can also be used to construct potential-based shaping rewards that provide dense intermediate learning signals without changing the underlying task objective, under the standard fixed-potential assumptions used in reward shaping.
Our use of this connection in the main paper is conceptual only: we do not perform RL fine-tuning with LenVM, and all reported results use LenVM at inference time.

\subsection{LenVM as a Value Function for a Length Objective}

Consider an autoregressive policy $\pi$ over a generated trajectory of length $L$.
Suppose the RL objective contains both a task component and a length component,
\begin{equation}
J(\pi) = J_{\mathrm{task}}(\pi) + s \, J_{\mathrm{len}}(\pi),
\end{equation}
where $s \in \mathbb{R}$ controls the strength and sign of the length term.
Following Section~\ref{sec:reward_return}, define the per-step length reward as
\begin{equation}
r_t^{\mathrm{len}} = -(1-\gamma), \qquad t=0,\dots,L-1,
\end{equation}
with terminal reward $r_L^{\mathrm{len}} = 0$.
The corresponding discounted return from state $s_t$ is then
\begin{equation}
G_t^{\mathrm{len}}
=
\sum_{i=0}^{L-t} \gamma^i r_{t+i}^{\mathrm{len}}
=
-(1-\gamma^{L-t}).
\end{equation}
Therefore the length value function under policy $\pi$ is
\begin{equation}
V_{\pi}^{\mathrm{len}}(s_t)
\triangleq
\mathbb{E}_{\pi}\!\left[G_t^{\mathrm{len}} \mid s_t\right].
\end{equation}
This is exactly the quantity LenVM is trained to estimate.
In this sense, LenVM is not merely correlated with generation length.
It is a value function for a well-defined token-level objective in which each additional decoding step incurs a constant discounted cost.

This interpretation is useful because it makes explicit what happens if LenVM is inserted directly into RL optimization.
Using a positive coefficient $s$ in the additive objective above means that the policy is optimized not only for task reward, but also for shorter continuations.
Thus, direct optimization of $J_{\mathrm{task}}(\pi) + s J_{\mathrm{len}}(\pi)$ generally changes the task objective unless generation efficiency is itself part of the intended goal.
This distinction is important.
LenVM can be used either as a genuine objective component, when one wishes to trade off quality against generation cost, or as a source of auxiliary structure for improving learning without redefining the task.
The latter case motivates the shaping perspective developed below.

\subsection{Advantage Decomposition and Policy Optimization}

With separate task and length rewards, one can also maintain separate value baselines.
Let $V^{\mathrm{task}}(s_t)$ denote the value function for the task reward and let $\hat{v}_{\phi}(s_t)$ denote the LenVM prediction for the length reward.
Then the temporal-difference residuals can be written as
\begin{equation}
\delta_t^{\mathrm{task}}
=
r_t^{\mathrm{task}}
+
\gamma_{\mathrm{task}} V^{\mathrm{task}}(s_{t+1})
-
V^{\mathrm{task}}(s_t),
\end{equation}
and
\begin{equation}
\delta_t^{\mathrm{len}}
=
r_t^{\mathrm{len}}
+
\gamma \hat{v}_{\phi}(s_{t+1})
-
\hat{v}_{\phi}(s_t).
\end{equation}
Applying GAE to each component gives
\begin{equation}
A_t^{\mathrm{task}}
=
\sum_{i=0}^{L-1-t}
(\gamma_{\mathrm{task}}\lambda_{\mathrm{task}})^i
\delta_{t+i}^{\mathrm{task}},
\end{equation}
and
\begin{equation}
A_t^{\mathrm{len}}
=
\sum_{i=0}^{L-1-t}
(\gamma\lambda_{\mathrm{len}})^i
\delta_{t+i}^{\mathrm{len}}.
\end{equation}
In the simplest case, one combines these signals additively:
\begin{equation}
A_t^{\mathrm{total}}
=
A_t^{\mathrm{task}} + s \, A_t^{\mathrm{len}}.
\end{equation}
One may also schedule the scalar coefficient $s$ over training if the intended quality-efficiency trade-off changes over time.
Here $s>0$ discourages longer continuations, while $s<0$ encourages them.

This decomposition is attractive because the quality and efficiency terms remain explicitly separated.
The task critic models the utility of the generated content, while LenVM models the expected future cost of continuing generation.
As a result, changing the scalar coefficient changes how these two factors enter the policy update without redefining the underlying value targets themselves.

At the same time, it is important to note that this additive formulation is an objective change, not merely a variance-reduction device.
When $s \neq 0$, the induced policy update explicitly prefers trajectories with different length characteristics.
Therefore this formulation is most appropriate when one truly wishes to optimize a performance-efficiency trade-off, such as token-budgeted generation, latency-aware decoding, or resource-constrained RL for language models.

\subsection{LenVM as a Potential for Policy-Invariant Reward Shaping}

The same learned signal also admits a different use that does not directly modify the underlying task objective.
Suppose the original RL problem is defined only by task reward $r_t^{\mathrm{task}}$.
If one wishes to provide dense intermediate feedback without changing the underlying task objective, a standard construction is potential-based reward shaping:
\begin{equation}
r_t'
=
r_t^{\mathrm{task}}
+
\beta \Bigl(\gamma \Phi(s_{t+1}) - \Phi(s_t)\Bigr),
\end{equation}
where $\Phi$ is any scalar potential over states and $\beta \ge 0$ controls the shaping strength.

LenVM provides a natural candidate for such a potential.
Because $\hat{v}_{\phi}(s_t) \in (-1,0)$ is a bounded monotone transform of expected remaining generation length, it defines a notion of prefix-wise progress toward termination.
States closer to EOS have values nearer to $0$, whereas states expected to continue longer have values closer to $-1$.
A simple choice is therefore
\begin{equation}
\Phi(s_t) = \mathrm{sg}\!\left(\hat{v}_{\phi}(s_t)\right),
\end{equation}
where $\mathrm{sg}(\cdot)$ denotes stop-gradient, and we set
\begin{equation}
\Phi(s_{\mathrm{EOS}})=0.
\end{equation}
The resulting shaping term is
\begin{equation}
F_t
=
\gamma \Phi(s_{t+1}) - \Phi(s_t).
\end{equation}

This construction clarifies the role of LenVM in RL.
If $\Phi(s_{t+1}) > \Phi(s_t)$, then the next decoding state is predicted to be closer to termination, so $F_t$ becomes larger.
Such a transition receives a more positive shaping signal.
Conversely, if $\Phi(s_{t+1}) < \Phi(s_t)$, then the chosen token moves the trajectory toward a longer continuation, and the shaping signal becomes smaller or more negative.
Thus LenVM-based shaping does not directly reward short outputs as an endpoint property.
Rather, it rewards local transitions that appear to make progress toward termination under the current decoding dynamics.

Under the standard shaping assumptions that the decoding process is treated as a discounted Markov decision process, the same discount factor is used in both the base return and the shaping term, and the potential is fixed during policy optimization, the reason this shaping can alter optimization dynamics without changing the underlying task objective is the telescoping form of the added reward.
For any trajectory starting from $s_0$,
\begin{equation}
\sum_{t=0}^{T} \gamma^t
\Bigl(\gamma \Phi(s_{t+1}) - \Phi(s_t)\Bigr)
=
-\Phi(s_0) + \gamma^{T+1}\Phi(s_{T+1}).
\end{equation}
When terminal states are assigned zero potential, this reduces to a boundary term depending only on the start state.
Therefore all trajectories from the same initial state receive the same offset, and their ordering under expected return is unchanged.
In this sense, potential-based shaping modifies the distribution of intermediate learning signals rather than the final task criterion itself.

This distinction is especially relevant for language-model RL.
A direct length penalty changes what the policy is asked to optimize.
By contrast, LenVM-based shaping can be interpreted as a progress signal that redistributes terminal information over token-level decisions.
It may therefore improve credit assignment and sample efficiency even when the intended objective is still purely task success.

In practice, however, LenVM is policy-dependent, since it estimates $V_{\pi}^{\mathrm{len}}(s_t)$ under the rollout policy.
For the shaping interpretation to remain closest to the classical policy-invariance result, it is preferable to use a pretrained and fixed LenVM when constructing $\Phi$.
If the potential is updated simultaneously with the policy, then the shaping reward itself becomes non-stationary, and the standard invariance guarantee no longer applies directly.
For this reason, if LenVM were to be used for RL shaping in future work, the most stable protocol would be to pretrain it first and then freeze it during policy optimization.

\subsection{Why This is Preferable to a Sequence-Level Length Penalty}

A common alternative is to add a sequence-level penalty such as $R_{\mathrm{task}} - \alpha L$ at the end of generation.
The token-level formulation above has several advantages.

First, it yields dense credit assignment.
Each additional token induces a local change in the length value, so the policy can receive an immediate signal about whether continuing the rollout appears worthwhile.
In contrast, a terminal penalty must be propagated backward through the full sequence.

Second, it admits a matched value baseline.
Since LenVM is trained directly on the return induced by $r_t^{\mathrm{len}}$, the resulting baseline is aligned with the optimization target for the length component.
This reduces variance relative to using a terminal length penalty with a generic value function that must jointly absorb task quality and sequence length.

Third, it cleanly separates quality from efficiency.
The task value network models the utility of the generated content, while LenVM models the expected future cost of additional decoding steps.
This decoupling is especially natural when task reward is sparse or delayed.

Fourth, in the shaping setting, LenVM provides a structured progress potential rather than only a final scalar penalty.
This is useful precisely because the reward is changed in a way that preserves the original optimum while making intermediate decisions easier to evaluate during training.

Finally, the trade-off coefficient remains interpretable.
In the additive objective formulation, $s$ acts directly on a length-specific advantage term.
In the shaping formulation, $\beta$ controls only the strength of the auxiliary progress signal.
These two roles should be distinguished, since one changes the optimized objective while the other changes only the learning dynamics.

\subsection{Connection to the Main Paper}

The role of this section is to clarify that the LenVM objective is compatible with PPO-style RL because it already defines a proper discounted value-learning problem over decoding states.
More importantly, the same learned signal admits two conceptually distinct uses in RL.
It can be treated as a length-specific value function and directly incorporated into the policy objective when one wishes to optimize quality-efficiency trade-offs.
Alternatively, it can be converted into a potential function for reward shaping under the standard fixed-potential assumptions when one wishes to preserve the original task objective while providing denser token-level learning signals.

This RL perspective also helps explain why LenVM is useful at inference time.
Even outside RL, the model provides a prefix-conditioned estimate of how close generation is to termination.
That signal can be exploited for length-controlled decoding and performance-efficiency trade-offs at test time.
We do not claim empirical RL gains in this paper, and we leave RL fine-tuning with these formulations to future work.

\begin{figure}[!t]
  \centering
  \includegraphics[width=\textwidth]{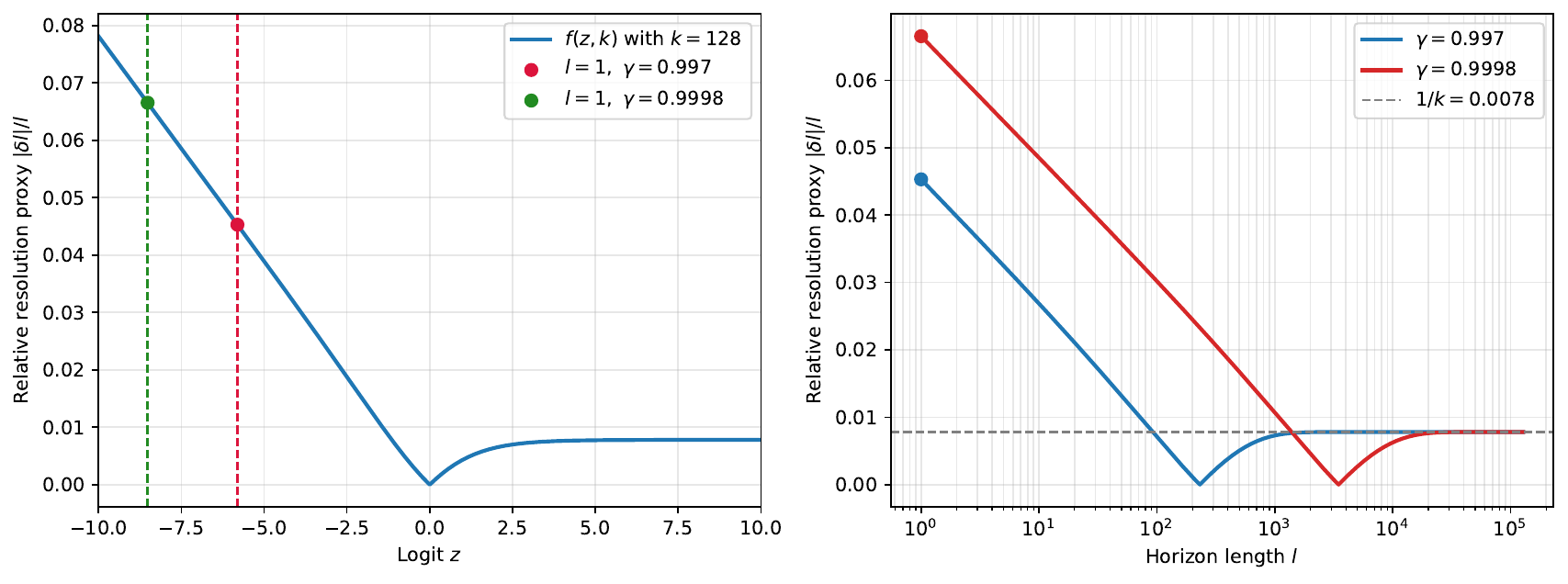}
  \caption{Relative length resolution under finite precision from two complementary views. Left: the proxy $f(z,k)$ as a function of the logit $z$. Right: the same quantity expressed as a function of the horizon $l$ using $\sigma(-z)=\gamma^l$. Markers indicate the finite-logit locations corresponding to $l=1$ for representative values of $\gamma$.}
  \label{figs:error_curve_combined}
\end{figure}

\section{Finite-Precision Analysis of Relative Length Resolution}
\label{app:precision_analysis_details}

LenVM is trained to regress remaining horizons over a wide dynamic range, from 1 to 32k. In this setting, the central numerical question is what relative length resolution the representation can maintain under finite precision. Since longer horizons naturally tolerate larger absolute deviations, the more relevant quantity is the relative perturbation $|\delta l|/l$.

Recall that LenVM predicts $\hat{v}=-\sigma(z)$ and that the length estimate is recovered from
\begin{equation}
1+\hat{v}=\gamma^{\hat{l}}.
\end{equation}
Since $1+\hat{v}=1-\sigma(z)=\sigma(-z)$, we have
\begin{equation}
\hat{l}=\frac{\ln(\sigma(-z))}{\ln\gamma},
\end{equation}
or equivalently,
\begin{equation}
\sigma(-z)=\gamma^l.
\end{equation}
Thus, the logit corresponding to horizon $l$ is
\begin{equation}
z(l)=\ln\frac{1-\gamma^l}{\gamma^l}.
\end{equation}

A first-order perturbation analysis with respect to $z$ gives
\begin{equation}
\left|\frac{\delta l}{l}\right|
\approx
m(z)\,|\delta z|,
\end{equation}
where
\begin{equation}
m(z)= -\frac{\sigma(z)}{\ln(\sigma(-z))}.
\end{equation}
To obtain a lightweight finite-precision proxy, we model the local logit perturbation as
\begin{equation}
|\delta z| \approx \frac{|z|}{k},
\end{equation}
where $k$ characterizes the effective precision level. In particular, taking $k \approx 2^p$ for a floating-point format with $p$ effective mantissa bits gives $k=128$ for bfloat16, $k=1024$ for fp16, and $k=8388608$ for fp32. Substituting this approximation into the expression above yields
\begin{equation}
\left|\frac{\delta l}{l}\right|
\approx
f(z,k)
=
-\frac{\sigma(z)\,|z|}{k\,\ln(\sigma(-z))}.
\end{equation}
We interpret $f(z,k)$ as a proxy for the local relative length resolution under finite precision.

For long horizons, corresponding to $z \to +\infty$, we have
\begin{equation}
\sigma(z)\to 1,
\qquad
\ln(\sigma(-z))\approx -z,
\end{equation}
and therefore
\begin{equation}
\left|\frac{\delta l}{l}\right|
\approx
\frac{1}{k}.
\end{equation}
Thus, at large horizons the representation approaches an approximately constant relative resolution floor. In contrast, the variation in relative resolution is concentrated in the short-horizon regime, where the corresponding logits remain finite and finite-precision effects are more pronounced. In practice, this means that the dominant numerical error arises in the small-length region rather than in the long-horizon tail.

Figure~\ref{figs:error_curve_combined} summarizes this analysis from two complementary views. The left panel shows the relative resolution proxy $f(z,k)$ directly in logit space, while the right panel rewrites the same quantity as a function of the horizon $l$ using $\sigma(-z)=\gamma^l$. These plots show that the relative error is primarily concentrated in the small-$l$ region, which is consistent with the empirical behavior observed in Figure~\ref{fig:gamma_ablation}. Together, they illustrate how finite-precision effects in logit space translate into horizon-dependent relative resolution.

\section{Future-Dependent Weighting Changes the Regression Target}
\label{app:future_dependent_weighting}

We clarify the population objective underlying LenVM training and show that weights depending on future rollout outcomes shift the regression target away from the state-conditional mean of the return proxy.

\paragraph{Setup.}
Let $\tau$ denote a sampled prompt-completion trajectory with generated length $L(\tau)$.
For each non-terminal step $t \in \{0,\dots,L(\tau)-1\}$, define the per-token squared error
\begin{equation}
    \ell_\theta(\tau,t) \triangleq \bigl(V_\theta(s_t^\tau) - G_t^\tau\bigr)^2,
    \qquad G_t^\tau = -(1-\gamma^{L(\tau)-t}).
\end{equation}

\subsection{General Weighted Objective}

Consider the general weighted population objective
\begin{equation}
    \mathcal{J}_{w}(\theta)
    = \frac{\mathbb{E}_{\tau,t}\!\left[\displaystyle\sum_{t=0}^{L(\tau)-1} w_t(\tau)\,\ell_\theta(\tau,t)\right]}
           {\mathbb{E}_{\tau,t}\!\left[\displaystyle\sum_{t=0}^{L(\tau)-1} w_t(\tau)\right]},
    \qquad w_t(\tau)\ge 0,
    \label{eq:appendix_weighted_objective}
\end{equation}
with empirical estimator
\begin{equation}
    \mathcal{L}_{w}
    = \frac{\displaystyle\sum_{n=1}^{N}\sum_{t=0}^{L^{(n)}-1} w_t(\tau^{(n)})\,\ell_\theta(\tau^{(n)},t)}
           {\displaystyle\sum_{n=1}^{N}\sum_{t=0}^{L^{(n)}-1} w_t(\tau^{(n)})}.
    \label{eq:appendix_emp_weighted_objective}
\end{equation}

\paragraph{Optimal predictor.}
Introducing the indicator $\mathbb{I}(s_t^\tau = s)$ and optimizing $V(s)$ pointwise, the first-order condition gives
\begin{equation}
    \mathbb{E}_{\tau,t}\!\left[\sum_{t=0}^{L(\tau)-1} w_t(\tau)\,\mathbb{I}(s_t^\tau = s)\cdot\bigl(V(s) - G_t^\tau\bigr)\right] = 0.
\end{equation}
Solving for $V(s)$ yields the minimizer of $\mathcal{J}_w$:
\begin{equation}
    V_w^*(s)
    = \frac{\mathbb{E}_{\tau,t}\!\left[w_t(\tau)\,G_t^\tau \mid s_t^\tau=s\right]}
           {\mathbb{E}_{\tau,t}\!\left[w_t(\tau)\mid s_t^\tau=s\right]},
    \label{eq:appendix_weighted_target}
\end{equation}
i.e., a $w$-weighted conditional mean of $G_t^\tau$ at state $s$.
Whether this coincides with the unweighted conditional mean $\mathbb{E}_{\tau,t}[G_t^\tau \mid s_t^\tau = s]$ depends on the structure of $w_t(\tau)$.

\subsection{Instantiations}

\paragraph{Case 1: Token-uniform weighting ($w_t(\tau) = 1$).}
Setting $w_t(\tau) = 1$ recovers the token-averaged objective
\begin{equation}
    \mathcal{J}_{\mathrm{tok}}(\theta)
    = \frac{\mathbb{E}_{\tau}\!\left[\displaystyle\sum_{t=0}^{L(\tau)-1}\ell_\theta(\tau,t)\right]}{\mathbb{E}_{\tau}[L(\tau)]},
    \label{eq:appendix_token_objective}
\end{equation}
where every non-terminal decoding step receives equal weight.
Since $w_t(\tau) = 1$ is constant, it cancels in Eq.~\eqref{eq:appendix_weighted_target}, and the optimal predictor is
\begin{equation}
    V_{\mathrm{tok}}^*(s) = \mathbb{E}_{\tau,t}\bigl[G_t^\tau \mid s_t^\tau = s\bigr].
    \label{eq:appendix_unweighted_target}
\end{equation}
This is the state-conditional mean of the return proxy, which is exactly the quantity LenVM aims to estimate.

\paragraph{Case 2: State-dependent weighting ($w_t(\tau) = c(s_t^\tau)$, $c(s) > 0$).}
If $w_t(\tau)$ depends only on the current state $s_t^\tau$, then conditioning on $s_t^\tau = s$ makes $w_t(\tau)$ a deterministic constant $c(s)$.
By the linearity of conditional expectation, $c(s)$ factors out of both the numerator and denominator in Eq.~\eqref{eq:appendix_weighted_target} and cancels:
\begin{equation}
    V_w^*(s)
    = \frac{c(s)\,\mathbb{E}_{\tau,t}\!\left[G_t^\tau \mid s_t^\tau=s\right]}
           {c(s)}
    = \mathbb{E}_{\tau,t}\bigl[G_t^\tau \mid s_t^\tau = s\bigr].
\end{equation}
Such weights only rebalance the relative importance of different states in the global objective; they do not alter the per-state regression target.

\paragraph{Case 3: Future-dependent weighting.}
If $w_t(\tau)$ depends on outcomes beyond $s_t^\tau$, then conditioned on $s_t^\tau = s$, the weight $w_t(\tau)$ remains a non-trivial random variable that may be correlated with $G_t^\tau$.
In general,
\begin{equation}
    V_w^*(s) \neq \mathbb{E}_{\tau,t}\bigl[G_t^\tau \mid s_t^\tau = s\bigr],
\end{equation}
meaning the regression target itself is shifted.
A concrete example is trajectory-level averaging, which sets $w_t(\tau) = 1/L(\tau)$:
\begin{equation}
    \mathcal{J}_{\mathrm{traj}}(\theta)
    = \mathbb{E}_{\tau}\!\left[\frac{1}{L(\tau)}\sum_{t=0}^{L(\tau)-1}\ell_\theta(\tau,t)\right].
    \label{eq:appendix_traj_objective}
\end{equation}
Because $L(\tau)$ is determined only after the full rollout, $1/L(\tau)$ is correlated with $G_t^\tau$ conditioned on $s_t^\tau = s$: trajectories with larger $L(\tau)$ receive smaller weight yet produce more negative $G_t^\tau$.
Consequently, $\mathcal{J}_{\mathrm{traj}}$ does not optimize the token-uniform objective in Eq.~\eqref{eq:appendix_token_objective}; it optimizes a future-length-reweighted variant with a shifted per-state target.

For LenVM, whose goal is dense token-level Monte Carlo regression over observed decoding states, we therefore adopt the token-averaged objective in Eq.~\eqref{eq:appendix_token_objective}.

\section{Why Inverting the Transformed Horizon Underestimates Expected Remaining Length}
\label{app:length_underestimation}

In Section~\ref{sec:first_token_pred}, we evaluate length prediction in the transformed space rather than by directly inverting the predicted horizon into a raw length estimate. This appendix explains why such inversion systematically underestimates the true expected remaining length.

Let $L$ denote the random remaining generation length from a given decoding state, with $L \ge 0$. Define the transformed horizon
\begin{equation}
u(L) = 1-\gamma^L,
\end{equation}
where $\gamma \in (0,1)$.

Suppose the model predicts the conditional expectation in this transformed space exactly:
\begin{equation}
\hat{u} = \mathbb{E}[u(L)].
\end{equation}
A natural raw length estimate is then obtained by inversion:
\begin{equation}
\hat{L} = u^{-1}(\hat{u}) = \frac{\ln(1-\hat{u})}{\ln\gamma}.
\end{equation}
We now show that this estimate is always less than or equal to the true expected remaining length $\mathbb{E}[L]$.

The first and second derivatives of $u(L)$ are
\begin{equation}
u'(L) = -\gamma^L \ln\gamma,
\end{equation}
and
\begin{equation}
u''(L) = -\gamma^L (\ln\gamma)^2.
\end{equation}
Because $\gamma \in (0,1)$, we have $\ln\gamma < 0$, so $u'(L) > 0$. Thus $u(L)$ is strictly increasing. Also, since $\gamma^L > 0$ and $(\ln\gamma)^2 > 0$, we have $u''(L) < 0$ for all $L$. Therefore, $u(L)$ is strictly concave.

By Jensen's inequality for concave functions,
\begin{equation}
\mathbb{E}[u(L)] \le u(\mathbb{E}[L]).
\label{eq:jensen_u}
\end{equation}
Since $u(\cdot)$ is strictly increasing, its inverse $u^{-1}(\cdot)$ is also strictly increasing. Applying $u^{-1}$ to both sides of Equation~\ref{eq:jensen_u} preserves the inequality:
\begin{equation}
u^{-1}(\mathbb{E}[u(L)]) \le u^{-1}(u(\mathbb{E}[L])) = \mathbb{E}[L].
\end{equation}
Therefore,
\begin{equation}
\hat{L} \le \mathbb{E}[L].
\end{equation}

As a result, the strict convexity of the discounted return mapping, combined with Jensen's inequality, guarantees that the inverted length $\hat{L}$ is bounded above by the true expected length $\mathbb{E}[L]$. The equality $\hat{L} = \mathbb{E}[L]$ holds if and only if the variance of $L$ is zero (i.e., the remaining length is completely deterministic). In all stochastic sequence generation scenarios where the remaining length has variance, the inversion will systematically underestimate the true expected length. In other words, even if the model predicts the conditional mean accurately in the transformed value space, directly mapping that prediction back to the raw length space via $u^{-1}$ yields an underestimate of the true expected remaining length. The underestimation issue is strictly a consequence of converting to the length space; there is no inherent problem when operating directly within the value space.

\section{Derivation of the Exponential Tilting Solution in the Performance–Efficiency Trade-off Experiment}
\label{app:exp_tilt_derivation}

We derive the closed-form solution to the KL-regularised length-steering objective in Eq.~\eqref{eq:kl_obj}.
At each decoding step $t$, let $p(x)$ be the base model's distribution over a finite candidate set $\mathcal{V}_t$ and let $\hat{v}(x)$ be the LenVM's value prediction for token $x$.
The objective has two goals: (1) minimise the expected LenVM value to steer generation toward shorter completions, and (2) penalise large deviations from the base model via KL divergence to preserve generation quality.
We seek the distribution $p'$ that solves, for $\beta < 0$:
\begin{equation}
  \min_{p'}\; \mathbb{E}_{p'}[\hat{v}(x)] - \frac{1}{\beta}\, D_{\mathrm{KL}}(p' \| p)
  \quad\text{subject to}\quad \sum_{x} p'(x) = 1,\; p'(x)\ge 0.
\end{equation}
Since $\beta < 0$, the term $-\frac{1}{\beta} > 0$, so the KL divergence enters with a positive coefficient, making the objective strictly convex in $p'$ and guaranteeing a unique global minimiser.
(If $\beta > 0$ were used instead, the KL term would be subtracted, rendering the objective unbounded below and the problem ill-posed.)

Expanding the KL term and introducing a Lagrange multiplier $\lambda$ for the normalisation constraint, the Lagrangian is
\begin{equation}
  \mathcal{L}(p', \lambda)
  = \sum_x p'(x)\,\hat{v}(x)
    - \frac{1}{\beta}\sum_x p'(x)\ln\frac{p'(x)}{p(x)}
    - \lambda\!\left(\sum_x p'(x) - 1\right).
\end{equation}

Taking the derivative with respect to $p'(x)$ and setting it to zero:
\begin{equation}
  \hat{v}(x) - \frac{1}{\beta}\!\left(\ln\frac{p'(x)}{p(x)} + 1\right) - \lambda = 0.
\end{equation}

Solving for $p'(x)$:
\begin{equation}
  p'(x) = p(x)\,\exp\!\bigl(\beta\,\hat{v}(x) - 1 - \beta\lambda\bigr).
\end{equation}

The normalisation condition $\sum_x p'(x)=1$ fixes the constant, giving the final result:
\begin{equation}
  p'(x) = \frac{p(x)\,\exp(\beta\cdot\hat{v}(x))}
               {\sum_{x'\in\mathcal{V}_t} p(x')\,\exp(\beta\cdot\hat{v}(x'))},
  \qquad \beta < 0.
\end{equation}

This is the Gibbs / softmax form, and it is the unique global minimiser.
With $\beta < 0$, tokens with lower predicted values (shorter expected horizons) receive higher probability mass, steering generation toward shorter completions while $|\beta|$ controls how far the distribution drifts from the base model.

\end{document}